\documentclass[10pt,twocolumn,letterpaper]{article}

\usepackage{cvpr}              
\usepackage{amsmath}
\usepackage{booktabs}
\usepackage{xcolor}
\usepackage{pifont}
\usepackage{tabularx}
\usepackage{booktabs}
\usepackage{xltabular}
\usepackage{array}

\providecommand{\cmark}{\ding{51}}
\providecommand{\xmark}{\ding{55}}
\providecommand{\best}[1]{\textbf{\textcolor{red}{#1}}}
\providecommand{\sbest}[1]{\textbf{\textcolor{blue}{#1}}}

\definecolor{cvprblue}{rgb}{0.21,0.49,0.74}
\usepackage[pagebackref,breaklinks,colorlinks,allcolors=cvprblue]{hyperref}

\def\confName{CVPR}
\def\confYear{2027}

\title{Towards a Multimodal Aesthetic Agent for Photographic Reference Grading}

\author{
Kunjie Lin$^{1}$ \quad
Haitao Zhang$^{1}$ \quad
Bosheng Chen$^{1}$ \quad
Lanhui Xu$^{1}$  \quad
Gaofeng Pan$^{1}$ \quad
Yuxin Lin$^{1}$ \\
Tianjiao Liu$^{2}$ \quad
Yi Yang$^{2,*}$ \quad
Yue Huang$^{1}$ \quad
Xinghao Ding$^{1,*}$ \quad
Zixu Lin$^{1,\dagger}$ \quad
Yunlong Lin$^{1,\dagger}$ \\[4pt]
$^{1}$Key Laboratory of Multimedia Trusted Perception and Efficient Computing, \\
Ministry of Education of China, 
Xiamen University, Xiamen, Fujian, China \\ \qquad
$^{2}$Central Media Technology Institute, Huawei \\
{\small $^{\dagger}$Project leaders. \qquad
$^{*}$Corresponding authors.}
}

\begin{document}
\maketitle
\begin{abstract}
Reference-guided photographic color grading helps users realize their aesthetic intent by selectively drawing on the color and tonal characteristics of a reference photograph.
Doing so requires understanding which characteristics users wish to adopt and translating these preferences into controllable edits.
However, image-only reference methods align overall appearance without instruction-grounded selection, while pixel-space generative editors typically produce flattened images without editable parameters, limiting subsequent refinement.
To address these challenges, we introduce \textbf{\textit{JarvisAura}}, a multimodal aesthetic agent that formulates photographic color grading as structured decision-making in an editable parameter space.
JarvisAura offers three key advantages:
(1) an aesthetic agent paradigm that integrates visual attribute perception, strategy-level reasoning, and parameterized execution for interpretable and controllable photographic color grading;
(2) a unified reasoning-to-execution pipeline that connects aesthetic planning with backend rendering and editable parameter refinement;
and (3) a three-stage training strategy comprising Parameter-Grounded Continual Pre-Training for parameter--appearance grounding, Reference-Grounded Supervised Fine-Tuning for reference-guided reasoning and parameter decisions, and Reference-Aligned Group Relative Policy Optimization for rendered-outcome refinement with multidimensional rewards and dynamic data derivation.
On Ref-Bench, JarvisAura achieves the best scores among methods with results across reported automatic metrics and a 74\% human-rated task-success rate, exceeding GPT-Image-2.5 by 9 percentage points.
\end{abstract}

\section{Introduction}
\label{sec:intro}

Photographic color grading helps users realize their aesthetic intent by shaping the color and atmosphere of their images.
Reference photographs provide concrete examples of color relationships, saturation, and tonal appearance~\cite{pitie2007linear, ke2023neural, larchenko2025color}.
Yet users may want to borrow only part of a reference's appearance, such as its warm palette but not its strong contrast.
Natural-language instructions can express these preferences.
We study reference-guided photographic color grading in which instructions determine which reference characteristics to adopt and how to adapt them to a source photograph.
The resulting adjustments should remain editable so that users can continue refining the image.

Existing methods support different parts of this process.
Image-only reference transfer methods use color distribution matching, statistical transformation, or color-space mapping~\cite{larchenko2025color, huang2023optimal, pouli2010progressive, pitie2007linear}, as well as deep feature transformation~\cite{wen2023cap, chiu2022photowct2, ho2021deep}, to change an image according to a reference.
They generally lack natural-language input, leaving users unable to specify which reference characteristics they want to adopt.
Generative image editors can use both reference images and textual instructions~\cite{team2023gemini, hurst2024gpt, wu2025qwen}, but pixel synthesis may introduce unintended content changes or local inconsistencies.
Their outputs also typically lack editing parameters, making further parameter-level adjustments difficult within a photographic post-processing workflow.
Text-driven parameterized editors produce editable operations from natural-language requests~\cite{lin2025jarvisart, lin2025jarvisevo, wu2026retouchiq}, but generally do not use a visual reference.
Users must therefore describe the desired appearance in words alone, which can be difficult when their preferences involve subtle color relationships or tonal effects.

\begin{figure*}[t]
    \centering
    \includegraphics[width=\linewidth]{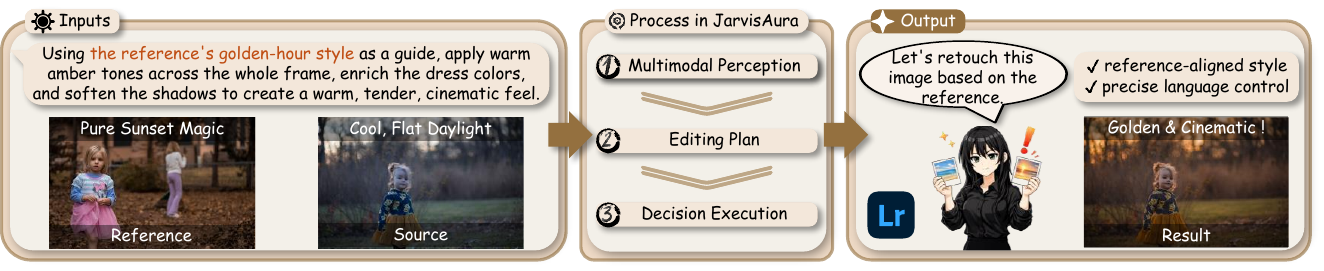}
\caption{
JarvisAura takes a source image, a reference image, and a user instruction, and produces a reference-driven color-grading result with an editable parameter representation. It identifies instruction-relevant grading attributes in the reference, determines which attributes to emphasize, and generates executable editing operations.
}
\label{fig:fig1}
\end{figure*}

We introduce \textbf{\textit{JarvisAura}}, a multimodal aesthetic agent for reference-guided photographic color grading.
Given a source image, a reference photograph, and a user instruction, JarvisAura reasons about which reference characteristics to adopt and which adjustments are needed to achieve them on the source.
It converts the grading plan into executable, editable parameters, as shown in Fig.~\ref{fig:fig1}.
Adobe Lightroom serves as a representative rendering backend.
The predicted parameters remain available for inspection and adjustment, allowing users to refine the rendered result.

We train JarvisAura in three stages.
Parameter-Grounded Continual Pre-Training (PG-CPT) establishes how grading parameters affect image appearance.
For this stage, we construct Ref-CPT with parameter-anchored descriptions of visual state changes.
Reference-Grounded Supervised Fine-Tuning (RG-SFT) then teaches the agent to choose parameters for a source--reference pair under a user instruction.
Its training dataset, Ref-SFT, provides expert reasoning trajectories and editing configurations that connect the intended appearance to specific grading decisions.

Reference-Aligned Group Relative Policy Optimization (RA-GRPO) further refines these decisions using their rendered outcomes.
We construct Ref-RL, an expert-guided reinforcement-learning dataset, and train with multidimensional outcome rewards.
This stage also addresses a difference between expert-derived training data and practical user requests.
In expert data, instructions and desired results are carefully aligned. In practice, users may express their aesthetic intent imprecisely or incompletely.
RA-GRPO therefore incorporates dynamic data derivation to create additional training examples from rendered images.
Acceptable rollouts that remain insufficiently aligned with the reference are reused as new source states and paired with expert-refined results.
New instructions are generated from the resulting image triplets with only visual cues as guidance, aiming to better approximate users’ imprecise or incomplete expressions of aesthetic intent.

On Ref-Bench, JarvisAura achieves the best performance on all reported automatic metrics among methods with available results.
Its human-rated task-success rate reaches 74\%, exceeding GPT-Image-2.5 by 9 percentage points.
Our contributions are summarized as follows:

\begin{itemize}[topsep=0pt, rightmargin=1em]
    \item We introduce JarvisAura, a multimodal aesthetic agent that uses reference photographs and user instructions to plan adjustments for source images and produce executable, editable grading parameters.

    \item We design two data-construction pipelines for Ref-CPT and Ref-SFT, providing parameter-anchored state-change descriptions and expert reference-guided reasoning trajectories with editing configurations.

    \item We develop a three-stage training strategy combining PG-CPT, RG-SFT, and RA-GRPO. The reinforcement-learning stage uses multidimensional outcome rewards and dynamic data derivation to refine rendered results and address the mismatch in intent clarity between expert annotations and real-world user expressions.
\end{itemize}

\section{Related Work}

\textbf{Reference-guided Photographic Color Grading.}
Existing reference-guided methods for photographic color grading mainly follow two directions: image-only transfer and multimodal generative editing. Image-only methods formulate transfer through color distribution matching, statistical transformation, color-space mapping~\cite{park2016efficient, huang2023optimal, larchenko2025color, pouli2010progressive, pitie2007linear}, or goal-conditioned optimization~\cite{wu2024goal}, while deep methods transform features or learn color-style correspondences~\cite{ho2021deep, wen2023cap, hong2021domain, ke2023neural, yoo2019photorealistic, chiu2022photowct2, li2017universal, li2018closed, an2020ultrafast}. 
Although they exploit references for global appearance adjustment, they provide limited language interaction and attribute-level control. Multimodal generative editors condition on multiple images and textual instructions~\cite{wu2025qwen, team2023gemini, hurst2024gpt}, with recent work further incorporating reasoning~\cite{yang2026seminft}. However, pixel-space synthesis may alter content, introduce local inconsistencies, and produce unstable results. These methods also typically output flattened images rather than executable parameters, hindering professional workflow integration and subsequent refinement. Thus, existing approaches do not jointly support visual references, instruction-guided emphasis, and editable parameterized outputs.

\begin{figure*}[t]
    \centering
    \includegraphics[width=\linewidth]{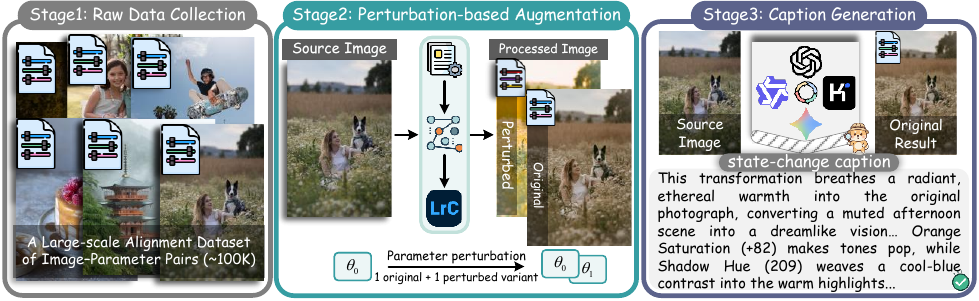} 
    \caption{
        Ref-CPT construction. The pipeline collects community edit pairs with configurations, renders results under original and perturbed settings, and generates parameter-anchored state-change captions for continual pre-training.
    }
\label{fig:data_pipeline4cpt}
\end{figure*} 

\textbf{Task-specific Agents.}
The development of multimodal foundation models has advanced task-specific agents in visual interaction, knowledge acquisition, and image processing. GUI agents execute software operations based on screen content and natural-language instructions~\cite{zhou2025mai, yan2025step, liu2025infiguiagent, wang2025ui, qin2025ui}, while knowledge acquisition and data analysis agents produce structured results through retrieval, reasoning, and tool use~\cite{zhang2025deepanalyze, narayan2025deepmmsearch, wu2025mmsearch}.
In image processing, recent studies have explored operation awareness~\cite{dutt2025monetgpt}, parameterized editing workflows~\cite{chen2025photoartagent}, natural-language-driven editable parameter generation~\cite{lin2025jarvisart}, reward-based editing optimization~\cite{wu2026retouchiq}, and editor--evaluator co-evolution~\cite{lin2025jarvisevo}. 
Such workflows connect high-level user intent with concrete editing operations, offering a foundation for more controllable and interpretable image manipulation.
However, existing methods mainly rely on textual instructions or predefined objectives, limiting their ability to exploit visual grading evidence from reference images. We address this gap by generating executable, parameterized grading results from multimodal references for photographic color grading.

\section{Method}
\subsection{Agent Formulation}
JarvisAura is a multimodal aesthetic agent for Reference-guided photographic reference Grading, as shown in Fig~\ref{fig:fig1}. Formally, the agent is defined as:
\begin{equation}
    (C, P) = F_{\theta}(I_{\mathrm{src}}, I_{\mathrm{ref}}, Q),
    \label{eq:overview}
\end{equation}
where $(I_{\mathrm{src}}, I_{\mathrm{ref}})$ denotes the source--reference image pair, $Q$ the user instruction, $C$ the grading analysis and plan, $P$ the executable parameter configuration, and $F_{\theta}$ JarvisAura.
The configuration is rendered by an editing backend as:
\begin{equation}
    I_{\mathrm{out}} = \mathcal{R}_{\mathcal{B}}(I_{\mathrm{src}}, P),
    \label{eq:rendering}
\end{equation}
where $I_{\mathrm{out}}$ denotes the rendered image and $\mathcal{R}_{\mathcal{B}}$ the renderer of backend $\mathcal{B}$. We instantiate $\mathcal{B}$ with Adobe Lightroom, while accommodating other backends by mapping $P$ to their parameter spaces or adapting the agent via data construction.

This formulation shows the limitations of existing methods. Image-only transfer methods ~\cite{larchenko2025color} omit $Q$, limiting control over user-specified attributes. Generative editors ~\cite{hurst2024gpt} directly synthesize $I_{\mathrm{out}}$, potentially altering source content, without providing editable parameters $P$ for subsequent refinement. Text-driven parameterized methods~\cite{lin2025jarvisart} accept $Q$ and predict $P$ but lack $I_{\mathrm{ref}}$, limiting visual grounding of target grading attributes.

\begin{figure*}[t]
    \centering
    \includegraphics[width=\linewidth]{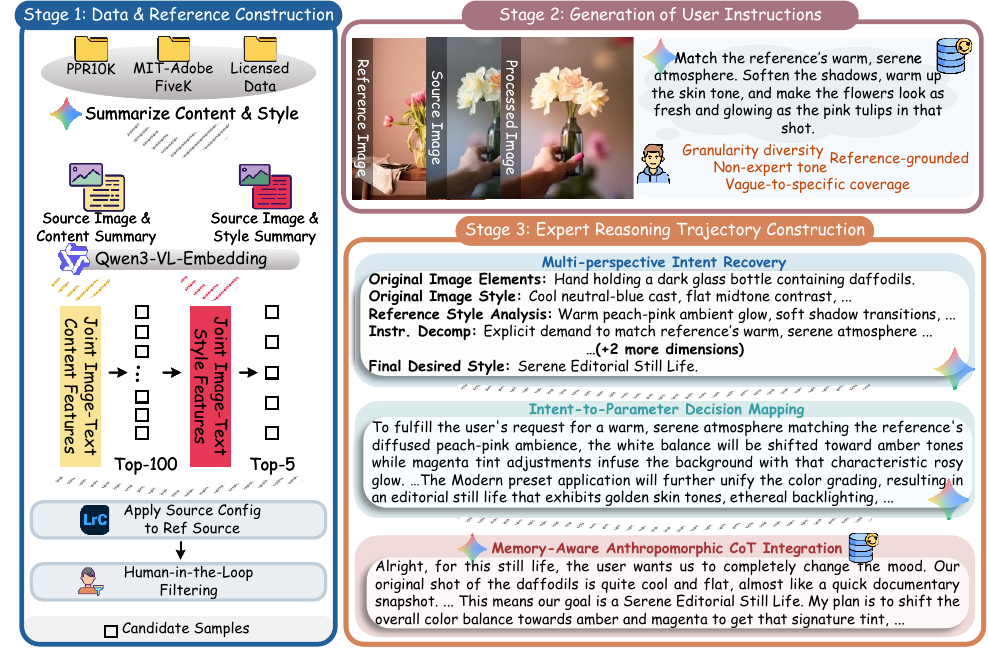} 
    \caption{
        Ref-SFT construction. The pipeline builds source--reference pairs, generates reference-grounded instructions, and synthesizes expert trajectories for planning and parameter decisions.
    }
    \label{fig:data_pipeline4sft}
\end{figure*} 

\subsection{Data Construction}

We build Ref-CPT and Ref-SFT through dedicated data-generation pipelines for continual pre-training and supervised fine-tuning, respectively. Ref-RL consists of samples refined by human experts, with data collection and quality-control details provided in Appendix~A.

\subsubsection{Data Generation for Continual Pre-training.} 
To ground parameter--appearance correspondences, Ref-CPT combines broad parameter-space coverage with detailed parameter-grounded state-change captions, as illustrated in Fig.~\ref{fig:data_pipeline4cpt}.

\textbf{Data Collection and Parameter Space Enhancement.} 
To obtain large-scale data, we collect approximately 100K image--configuration pairs from creator communities, in compliance with their terms of use and for academic research only.
To scale up samples and broaden parameter-space coverage, we perturb editing parameters and clamp them to valid ranges, then render results under both original and perturbed settings via an open-source protocol \cite{lin2025jarvisart}.
Through this augmentation, we construct about 200K high-quality triplets of original images, parameterized configurations, and resulting images.
This scale sufficiently covers the parameter space, enabling the dataset to capture how parameter values affect appearance in trend and relative strength, and supporting the model’s learning of professional parametric retouching knowledge.

\textbf{Automated State-Transition Captioning.} With large-scale sample pairs, we use advanced MLLMs for automated annotation. To generate professionally grounded, information-dense captions, we design concise prompts (see Appendix A) that jointly analyze the original image, parameterized configuration, and the rendered result.
Unlike generic semantic image descriptions, Ref-CPT annotations focus on a fine-grained characterization of the retouching trajectory.
To improve annotation quality and diversity, we use four multimodal models, Kimi-K2.5 \cite{team2026kimi}, Qwen3-VL-235 \cite{bai2025qwen3}, Gemini-3-Pro \cite{team2023gemini}, and GPT-5.2 \cite{hurst2024gpt}, with Qwen3-VL-235 from the same family as our backbone.
We rotate among these models when generating captions and enforce a unified prompting scheme for output format and terminology, yielding high-quality, diverse state-change captions. To ensure annotation consistency and reduce supervision mismatch, we use a shared VLM-as-Judge verifier to assess caption validity and faithfulness to the underlying edits, with more details provided in Appendix~A.

\subsubsection{Data Generation for Supervised Fine-tuning.}
Following Fig.~\ref{fig:data_pipeline4sft}, we construct Ref-SFT to model expert-style reasoning and parameter decisions for SFT.
Using the notation above, each sample is represented as $\langle I_{\text{src}}, I_{\text{ref}}, I_{\mathrm{out}}, Q, C, P \rangle$.

\textbf{Data Generation and Reference Construction.} 
This stage builds visually informative reference relations from large-scale data.  
We integrate MIT-Adobe FiveK \cite{bychkovsky2011learning}, PPR10K \cite{liang2021ppr10k}, other licensed sources, obtaining image pairs together with their parameterized configurations.
To ensure content match and style transferability, we use a two-stage retrieval strategy: 
1) \textbf{In semantic retrieval}, we generate content summaries with Gemini-3-Pro and use Qwen3-VL-Embedding \cite{li2026qwen3} to encode fused image-summary representations, retrieving the top 100 candidates.
2) \textbf{In style reranking}, we summarize candidates’ style attributes and compute similarity using image-style embeddings, selecting the top 5 as reference originals.
Finally, we apply the source parameterized configuration to the selected reference originals to obtain the reference image $I_{\text{ref}}$. 
To incorporate human aesthetic judgment and ensure data quality, we apply human-in-the-loop screening to filter mismatched or low-quality samples, completing $\langle I_{\text{src}}, I_{\text{ref}}, I_{\text{tgt}}, P \rangle$.

\textbf{User-Instruction Generation.}
In this stage, we use Gemini-3.1-Pro \cite{team2023gemini} with role-playing prompting to elicit user-like instructions.
To promote expression diversity, we add a simple sliding-window memory that discourages repetitive phrasing when generating reference-grounded instructions.
These instructions vary in granularity, wording, and emphasis, covering interactions from vague aesthetic requests to specific functional edits.
Prompt templates are provided in Appendix~A.
In this step, we map $\langle I_{\text{src}}, I_{\text{ref}}, I_{\text{tgt}}, P \rangle$ to $\langle I_{\text{src}}, I_{\text{ref}}, I_{\text{tgt}}, Q, P \rangle$.

\textbf{Expert Reasoning Trajectory Construction.} This key stage aims to capture expert reasoning via progressive prompting.
First, the model decomposes the task using cues such as source content, reference style, and user instruction to infer the underlying intent.
Then, using the parameterized configuration, we map the inferred intent into structured decision logic that specifies parameter edits and expected visual effects, linking numeric adjustments to aesthetic goals.
Finally, role-playing prompts integrate the intent analysis and decision logic into human-like reasoning trajectories.
We introduce a memory mechanism, similar to instruction generation, to avoid templated outputs and keep CoTs structured while diversifying colloquial phrasing.
These steps yield approximately 50K CoT-supervised Ref-SFT samples of the form $\langle I_{\text{src}}, I_{\text{ref}}, I_{\mathrm{out}}, Q, C, P \rangle$.

\begin{figure*}[t]
    \centering
    \includegraphics[width=\linewidth]{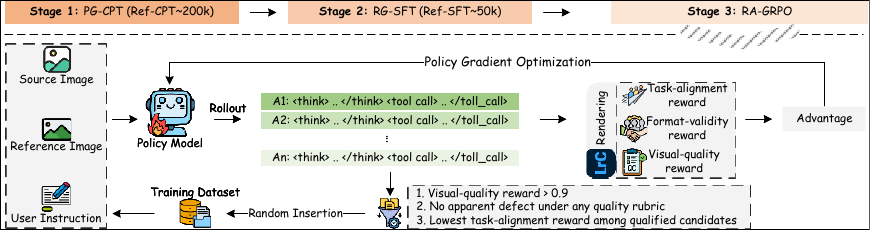}
    \caption{
        JarvisAura post-training framework: we first perform PG-CPT to ground color-grading parameter knowledge, then conduct RG-SFT to learn reference-guided reasoning and parameter decisions, and finally apply RA-GRPO with multidimensional rewards and dynamic data derivation to improve parameter accuracy and alignment with reference images and user instructions.
    }
    \label{fig:train}
\end{figure*} 

\subsection{Training Strategy}

\subsubsection{Parameter-Grounded Continual Pre-training.}
Pre-training corpora for general-purpose multimodal models underrepresent professional photographic color grading, leaving them without reliable parameter--appearance correspondences. Following the \textit{Superficial Alignment Hypothesis} \cite{zhou2023lima, zhuang2025hephaestus}, we inject this knowledge through Parameter-Grounded Continual Pre-training on Ref-CPT.
State-change captions provide explicit supervision by describing appearance changes induced by parameterized adjustments.
This enables the model to learn: 1) visual effects of each parameter across different values; 2) parameter interactions under joint parameter adjustment; and 3) content-dependent variation in parameter effects.

\subsubsection{Reference-Grounded Supervised Fine-tuning.}
Parameter knowledge alone cannot turn the source image and multimodal reference into actionable grading decisions. We therefore perform Reference-Grounded Supervised Fine-Tuning (RG-SFT) on Ref-SFT, supervising reasoning trajectories and executable parameters to learn reference-guided decisions.

\subsubsection{Reference-Aligned Group Relative Policy Optimization.}

Improving parameter accuracy and reducing instruction mismatch with limited expert annotation remain challenging. Building on CPT and SFT, we propose RA-GRPO, which combines multidimensional outcome rewards with dynamic data derivation during training, as shown in the lower half of Fig.~\ref{fig:train}.

\textbf{Reinforcement Learning for Accurate Decisions.}
User evaluation is anchored to the reference image, placing stringent demands on parameter accuracy. SFT primarily fits demonstrated decisions and does not directly optimize them through rendering feedback. We therefore apply Group Relative Policy Optimization (GRPO) to optimize candidate outputs sampled for each multimodal input. Given an input $x$, we sample $G$ candidate outputs $\{y_i\}_{i=1}^{G}$ from the old policy $\pi_{\theta_{\mathrm{old}}}$ and compute rewards from their rendered results. The objective is defined as:

\begin{equation}
\ell_{i,t}
=
\min\bigl\{
    \rho_{i,t}A_i,\,
    \operatorname{clip}(\rho_{i,t},1-\epsilon,1+\epsilon)A_i
\bigr\}.
\label{eq:token_objective}
\end{equation}

\begin{equation}
\mathcal{J}_{\mathrm{GRPO}}(\theta)
=
\mathbb{E}\left[
    \frac{1}{G}\sum_{i=1}^{G}
    \frac{1}{|y_i|}\sum_{t=1}^{|y_i|}
    \ell_{i,t}
\right].
\label{eq:grpo}
\end{equation}

where
$\rho_{i,t}(\theta)=
\frac{\pi_{\theta}(y_{i,t}\mid x,y_{i,<t})}
     {\pi_{\theta_{\mathrm{old}}}(y_{i,t}\mid x,y_{i,<t})}$
is the token-level probability ratio. $\pi_{\theta}$ denote the current policies, $y_{i,t}$ is the $t$-th token of candidate $y_i$, and $y_{i,<t}$ denotes its preceding tokens. The clipping threshold is denoted by $\epsilon$, and $A_i$ is the relative advantage computed from the group rewards:

\begin{equation}
    A_i=
    \frac{R_i-\operatorname{mean}(\{R_j\}_{j=1}^{G})}
    {\operatorname{std}(\{R_j\}_{j=1}^{G})+\epsilon}.
    \label{eq:grpo_advantage}
\end{equation}

The reward comprises three components, measuring task alignment, visual quality, and output validity, respectively, and is defined as 
$R_i=R_i^{\mathrm{task}}R_i^{\mathrm{qual}}R_i^{\mathrm{fmt}}$.

\textbf{\textit{1) Task-alignment reward.}} It combines reference matching with instruction following:

\begin{equation}
\begin{aligned}
R_i^{\mathrm{task}}
&=
\underbrace{
\operatorname{RankScore}\!\left(
d\!\left(I_i^{\mathrm{out}},I_{\mathrm{tgt}}\right);
\left\{
d\!\left(I_j^{\mathrm{out}},I_{\mathrm{tgt}}\right)
\right\}_{j=1}^{G}
\right)
}_{\text{reference matching}}
\\
&\quad+
\underbrace{
\mathcal{V}_{\mathrm{inst}}\!\left(
I_{\mathrm{src}},I_{\mathrm{ref}},I_i^{\mathrm{out}},Q,P_{i}
\right)
}_{\text{instruction following}}.
\end{aligned}
\label{eq:task_reward}
\end{equation}

where $I_i^{\mathrm{out}}$ and $P_i$ denote candidate $i$'s rendered result and predicted configuration, respectively, and $I_{\mathrm{tgt}}$ the target. The distance $d(\cdot,\cdot)$ is the Earth Mover's Distance between normalized Lab $a$--$b$ color histograms. $\operatorname{RankScore}(\cdot)$ maps distances to within-group scores, assigning higher scores to smaller distances. $\mathcal{V}_{\mathrm{inst}}(\cdot)$ provides parameter-aware instruction feedback, grounding evaluation in executable decisions and mitigating hallucinations. Detailed protocols appear in Appendix~B.

\textbf{\textit{2) Visual-quality reward.}} It evaluates the intrinsic visual quality:

\begin{equation}
    R_i^{\mathrm{qual}}
    =Q_i
    =\frac{1}{100K}\sum_{k=1}^{K}q_{i,k},
    \label{eq:quality_reward}
\end{equation}

where $q_{i,k}$ is the score for the $k$-th quality criterion and $K=14$ is the total number of criteria. The rubric uses four score levels, with their definitions and aggregation detailed in Appendix~B.

\textbf{\textit{3) Format-validity reward.}}
To encourage structured and executable outputs, $R_i^{\mathrm{fmt}}$ evaluates output-format compliance and parameter validity. At the response level, it checks the structure of \texttt{<think>...</think>} and \texttt{<tool\_call>}$P$\texttt{</tool\_call>} blocks and the parseability of configuration $P$. At the parameter level, it verifies that parameter names are valid and values fall within their permitted ranges. Detailed checks and constraints are provided in Appendix~B.

\textbf{Dynamic Data Derivation.} We use human-annotated expert samples for reinforcement learning, with the detailed construction procedure provided in the Appendix~A.
These rigorously validated samples establish well-aligned correspondences between user instructions and desired results. In real-world use, however, users may express their aesthetic intent imprecisely or incompletely, leaving some preferences implicit.
To bridge this gap, we derive training samples from rollout candidates. We retain candidates whose visual-quality reward satisfies $R_i^{\mathrm{qual}}>0.95$ and that show no apparent defect under any quality rubric. Among them, we select the candidate with the lowest task-alignment reward. 
Its rendered result is treated as a new source state and paired with the reference image and expert-refined result to form an image triplet. 
We prompt Gemini-3.1-Pro to simulate user instructions from visual information alone, excluding editing parameters, to better approximate users’ potentially vague or imprecise expressions of aesthetic intent in real-world interactions. The prompt template is provided in Appendix A.
The derived sample is randomly added to the training pool.

\begin{table*}[t]
\centering
\caption{Quantitative comparison on Ref-Bench (300 samples). We highlight the \best{best} and \sbest{second-best} results. SC, PQ, and O are evaluated by Gemini-3.1-Pro. Methods in the Ref.-Transfer group do not accept textual instructions.}
\label{tab:main_comparison}

\setlength{\tabcolsep}{4.5pt}
\renewcommand{\arraystretch}{1.05}

\begin{tabular}{@{}lcccccccc@{}}
\toprule
Method & Instr.
& $\mathrm{L1}_{\times 10^{2}}$ $\downarrow$
& $\mathrm{L2}_{\times 10^{3}}$ $\downarrow$
& CIEDE2000 $\downarrow$
& SC $\uparrow$ & PQ $\uparrow$ & O $\uparrow$
& HTSR $\uparrow$ \\
\midrule
ModFlows~\cite{larchenko2025color}
& \xmark & 11.86 & \sbest{16.89} & \sbest{11.16}
& -- & -- & -- & -- \\
RLPixTuner~\cite{wu2024goal}
& \xmark & 17.16 & 51.46 & 18.84
& -- & -- & -- & -- \\
\midrule
UniPic-3~\cite{wang2025skywork}
& \cmark & 22.12 & 89.15 & 23.61
& 4.87 & 3.86 & 4.21 & 0.09 \\
FireRed-Image-Edit~\cite{firered2026rededit}
& \cmark & 12.73 & 33.02 & 14.17
& 5.95 & 5.97 & 5.83 & 0.43 \\
Qwen-Image-2.1~\cite{wu2025qwenimagetechnicalreport}
& \cmark & 13.60 & 36.76 & 15.03
& 5.89 & 6.20 & 5.91 & 0.38 \\
JoyAI-Image~\cite{song2026joyai}
& \cmark & 18.87 & 70.40 & 20.23
& 5.23 & 4.00 & 4.41 & 0.15 \\
\midrule
Nano-Banana-2~\cite{team2023gemini}
& \cmark & 12.59 & 32.07 & 13.90
& 6.20 & 6.96 & 6.49 & 0.55 \\
GPT-Image-2.5~\cite{hurst2024gpt}
& \cmark & \sbest{10.23} & 20.47 & 11.67
& \sbest{6.72} & 7.18 & \sbest{6.90} & \sbest{0.65} \\
Seedream-5-pro~\cite{seedream2025seedream}
& \cmark & 10.97 & 24.04 & 12.37
& 6.43 & \sbest{7.21} & 6.75 & 0.58 \\
\midrule
\textbf{JarvisAura}
& \cmark & \best{8.92} & \best{13.02} & \best{10.09}
& \best{6.97} & \best{7.51} & \best{7.19} & \best{0.74} \\
\bottomrule
\end{tabular}
\end{table*}

\begin{table*}[t]
\centering
\caption{A/B preference ratio for JarvisAura on cases where both JarvisAura and the compared baseline produce successful outputs. The ratio is defined as $(N_{\mathrm{win}}+N_{\mathrm{tie}})/(N_{\mathrm{loss}}+N_{\mathrm{tie}})$.}
\label{tab:ab_preference}

\setlength{\tabcolsep}{4.5pt}
\renewcommand{\arraystretch}{1.05}

\begin{tabular}{@{}lcccc@{}}
\toprule
 & GPT-Image-2.5 & Seedream-5-pro
 & Nano-Banana-2 & FireRed-Image-Edit \\
\midrule
Preference ratio for ours $\uparrow$
 & 1.12 & 1.30 & 1.47 & 1.53 \\
\bottomrule
\end{tabular}
\end{table*}

\section{Experiment}
\subsection{Experimental Setup}
\textbf{Implementation details.} We develop JarvisAura through three-stage post-training of Qwen3-VL-8B-Instruct~\cite{bai2025qwen3}.
(1) For PG-CPT, we train on Ref-CPT ($\sim$200K state-change captions) for one epoch using ms-swift~\cite{zhao2024swiftascalablelightweightinfrastructure}, with batch size 1 and learning rate $5\times10^{-5}$.
(2) For RG-SFT, we fine-tune the PG-CPT checkpoint on Ref-SFT ($\sim$50K CoT-supervised samples) for two epochs using LLaMAFactory~\cite{zheng2024llamafactory}, with batch size 1 and learning rate $1\times10^{-5}$.
(3) For RA-GRPO, we train on Ref-RL ($\sim$2K samples) using verl~\cite{sheng2025hybridflow} with batch size 8, learning rate $5\times10^{-6}$, and 8 rollouts per input. Further details are provided in Appendix~C.

\noindent\textbf{Evaluation Setup.}
Ref-Bench contains 300 test instances with consensus-validated, expert-edited ground-truth targets, covering common photographic scenes and disjoint from the training data.
We measure fidelity to expert targets using L1/L2 and CIEDE2000 ($\Delta E_{00}$). Following prior evaluation protocols~\cite{lin2025jarvisart, zhang2023magicbrush}, Gemini-3.1-Pro~\cite{team2023gemini} scores Semantic Consistency (SC) and Perceptual Quality (PQ) on a 0--10 scale, with $O=\sqrt{\mathrm{SC}\times\mathrm{PQ}}$.
With expert targets withheld, a blinded user study measures Human-rated Task Success Rate (HTSR) for JarvisAura and input-compatible generative baselines based on fine-grained instruction fulfillment, alignment with requested reference attributes, visual quality, and content preservation.
More details are provided in Appendix~C.

\noindent\textbf{Baselines.}
We compare JarvisAura with two categories of baselines.
Image-only transfer methods, including ModFlows~\cite{larchenko2025color} and RLPixTuner~\cite{wu2024goal}, receive only source--reference image pairs.
Multimodal generative editors additionally receive user instructions and include UniPic-3~\cite{wang2025skywork}, FireRed-Image-Edit~\cite{firered2026rededit}, Qwen-Image-2.1~\cite{wu2025qwenimagetechnicalreport}, JoyAI-Image~\cite{song2026joyai}, Nano-Banana-2~\cite{team2023gemini}, GPT-Image-2.5~\cite{hurst2024gpt}, and Seedream-5-pro~\cite{seedream2025seedream}.

\subsection{Experimental Results}
\noindent\textbf{Quantitative comparison.}
As shown in Table~\ref{tab:main_comparison}, JarvisAura achieves the best results across all reported automatic metrics. Lower L1, L2, and CIEDE2000 values indicate closer agreement with consensus-validated, expert-refined targets in terms of pixel-level error and perceptual color difference. Relative to ModFlows, JarvisAura reduces L2 and CIEDE2000 by approximately 22.9\% and 9.6\%, respectively. Compared with GPT-Image-2.5 under the same multimodal inputs, it reduces CIEDE2000 by approximately 13.5\% while achieving higher SC (6.97 vs.\ 6.72) and O (7.19 vs.\ 6.90). Its PQ of 7.51 also exceeds the second-best score of 7.21 from Seedream-5-pro. These results demonstrate stronger target fidelity alongside improved semantic consistency and perceptual quality.

\begin{figure*}[!t]
    \setlength{\abovecaptionskip}{0.1cm}
    \setlength{\belowcaptionskip}{-0.5cm}
    \centering
    \includegraphics[width=0.95\linewidth]{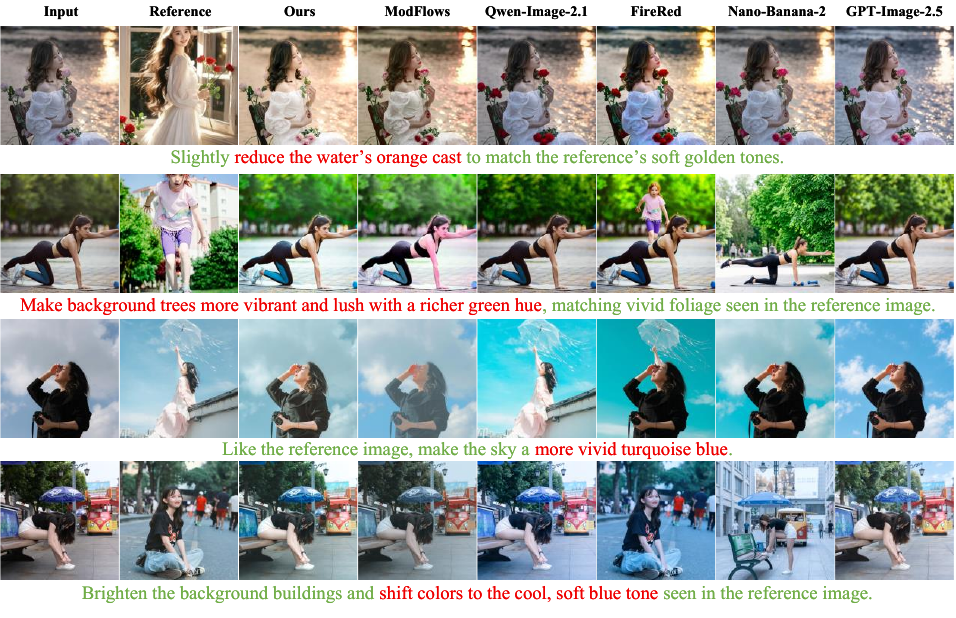}
    \caption{Visual comparison of different methods on Ref-Bench. JarvisAura produces faithful retouching results that closely match the reference style while preserving source content.
    }
    \label{fig:visual_results}
\end{figure*}

\noindent\textbf{Qualitative comparison.}
As shown in Fig.~\ref{fig:visual_results}, image-only transfer methods only perform simple color transfer and lack textual conditioning to capture users’ emphasis on reference attributes. Some generative outputs blend source and reference content and deviate from the requested color attributes. In contrast, JarvisAura more accurately transfers instruction-specified reference attributes while preserving source content.

\begin{table*}[t]
    \centering
    \caption{Ablation studies on Ref-Bench.}
    \label{tab:ablation}

    \setlength{\tabcolsep}{4.5pt}
    \renewcommand{\arraystretch}{1.0}

    \begin{tabular}{@{}lcccccc@{}}
        \toprule
        \textbf{Configurations}
        & $\mathrm{L1}_{\times 10^{2}}$ $\downarrow$
        & $\mathrm{L2}_{\times 10^{3}}$ $\downarrow$
        & CIEDE2000 $\downarrow$
        & SC $\uparrow$ & PQ $\uparrow$ & O $\uparrow$ \\
        \midrule
        \multicolumn{7}{l}{\textcolor{gray!60}{Training stages}} \\
        RG-SFT only
        & 11.04 & 25.00 & 12.46 & 5.60 & 6.83 & 6.16 \\
        PG-CPT + RG-SFT
        & 10.25 & 21.59 & 11.77 & 6.13 & 7.02 & 6.53 \\
        PG-CPT + RG-SFT + RA-GRPO (Ours)
        & 8.92 & 13.02 & 10.09 & 6.97 & 7.51 & 7.19 \\
        \midrule
        \multicolumn{7}{l}{\textcolor{gray!60}{Dynamic data derivation}} \\
        Derived data only
        & 9.60 & 16.64 & 10.46 & 6.59 & 7.24 & 6.88 \\
        w/o data derivation
        & 9.55 & 17.45 & 10.34 & 6.68 & 7.31 & 6.96 \\
        Original + derived data (Ours)
        & 8.92 & 13.02 & 10.09 & 6.97 & 7.51 & 7.19 \\
        \bottomrule
    \end{tabular}
\end{table*}

\noindent\textbf{Human evaluation.}
To account for subjective aesthetic preferences, we assess task success and pairwise preference through a blinded user study, with expert targets withheld. JarvisAura achieves an HTSR of 74\%, exceeding the strongest baseline, GPT-Image-2.5, by 9 percentage points (Table~\ref{tab:main_comparison}). For each baseline, we additionally conduct A/B evaluation on cases where both methods produce successful outputs. Preference ratios range from 1.12 to 1.53, favoring JarvisAura across all four comparisons (Table~\ref{tab:ab_preference}). These findings complement automatic metrics with human judgments of task success and relative preference. Detailed protocols are provided in Appendix~C.

\section{Ablation Study}
\textbf{Post-training stages.}
We compare RG-SFT alone, PG-CPT followed by RG-SFT, and the complete three-stage pipeline to examine the role of each training stage. RG-SFT establishes reference-grounded reasoning and parameter supervision. PG-CPT addresses the parameter--appearance correspondences required for accurate color and tone adjustment. Its inclusion reduces CIEDE2000 from 12.46 to 11.77 and raises $O$ from 6.16 to 6.53. 
Relative to the PG-CPT and RG-SFT configuration, the complete pipeline further reduces L2 from 21.59 to 13.02 and CIEDE2000 from 11.77 to 10.09, while increasing $O$ from 6.53 to 7.19. The concurrent gains in target fidelity and overall semantic--perceptual quality indicate that outcome-level policy optimization complements supervised parameter prediction by refining decisions according to their rendered results.

\textbf{Dynamic data derivation.}
Compared with training without dynamic data derivation, training on both original and derived data reduces L2 from 17.45 to 13.02 and raises $O$ from 6.96 to 7.19, with consistent improvements in CIEDE2000, SC, and PQ. This result indicates that derived samples provide complementary training signal alongside the original parameter-grounded data. 
To isolate the standalone contribution of derivation, we construct a derived-only control from the collected derived samples, supplementing them with SFT-rollout samples under the same derivation procedure to match the size of the no-derivation training set. 
As shown in Table~\ref{tab:ablation}, comparable performance to the no-derivation control suggests that the derived samples retain meaningful training value despite excluding 
the original data.

\section{Conclusion}
We present JarvisAura, a multimodal reference agent that maps a source--reference pair and a user instruction to executable, editable parameters for photographic color grading. We construct Ref-CPT with approximately 200K parameter-grounded state-change captions, Ref-SFT with approximately 50K reference-grounded CoT samples, and Ref-RL with expert-guided reinforcement-learning samples. Its training pipeline comprises three stages: (1) PG-CPT grounds parameter--appearance correspondences, (2) RG-SFT learns reference-grounded reasoning and parameter decisions, and (3) RA-GRPO optimizes rendered outcomes through multidimensional rewards and dynamic data derivation. On the benchmark, JarvisAura achieves the best performance on all reported automatic metrics. In the blinded human evaluation, it attains the highest task-success rate among the evaluated input-compatible methods. Pairwise comparisons with four selected leading generative baselines favor JarvisAura on jointly successful cases. Ablation results further support the effectiveness of the overall approach. Collectively, these findings support JarvisAura as an effective approach to multimodal, reference-guided photographic color grading.

{
    
    \small
    \bibliographystyle{ieeenat_fullname}
    \bibliography{main}
}
\clearpage
\appendix

\setcounter{figure}{0}
\setcounter{table}{0}
\setcounter{equation}{0}
\renewcommand{\thefigure}{S\arabic{figure}}
\renewcommand{\thetable}{S\arabic{table}}
\renewcommand{\theequation}{S\arabic{equation}}
\renewcommand{\theHfigure}{appendix.\arabic{figure}}
\renewcommand{\theHtable}{appendix.\arabic{table}}
\renewcommand{\theHequation}{appendix.\arabic{equation}}

\newcommand{\suppfigure}[3]{
  \begin{figure}[tbp]
    \centering
    \IfFileExists{#1}{
      \includegraphics[width=\linewidth,height=0.8\textheight,keepaspectratio]{#1}
    }{
      \fbox{\parbox[c][0.1\textheight][c]{0.92\linewidth}{
        \mbox{}
      }}
    }
    \caption{#2}
    \label{#3}
  \end{figure}
}

\section*{Appendices}

The appendices are organized into four sections covering dataset details in Section~\ref{sec:supp-data}, methodological details in Section~\ref{sec:supp-methodology}, additional experimental details in Section~\ref{sec:supp-exp-details}, and additional qualitative results in Section~\ref{sec:supp-qualitative}.

\begin{itemize}
  \item \textbf{Section~\ref{sec:supp-data}: Dataset Details.}
  Statistics and composition, comparison with existing datasets, data construction, representative samples, and data-generation prompt figures.
  \item \textbf{Section~\ref{sec:supp-methodology}: Methodological Details.}
  Data quality control, task-alignment rewards, visual-quality rewards, format-validity rewards, and dynamic data derivation.
  \item \textbf{Section~\ref{sec:supp-exp-details}: Additional Experimental Details.}
  Training and baseline configurations, Ref-Bench construction, automatic evaluation, human evaluation, and computational cost.
  \item \textbf{Section~\ref{sec:supp-qualitative}: Additional Qualitative Results.}
  Additional visual comparisons and intermediate reasoning and parameter outputs.
\end{itemize}

\section{Dataset Details}
\label{sec:supp-data}

\subsection{Statistics and Composition}
\label{sec:supp-data-statistics}

RefGrade is a multimodal dataset for Instruction-Selective Photographic Reference Grading, organized into three complementary components. Ref-CPT provides approximately 200K parameter-grounded state-change captions for Parameter-Grounded Continual Pre-Training. Ref-SFT contains approximately 50K reference-grounded samples pairing user instructions with reasoning trajectories and parameter configurations for Reference-Grounded Supervised Fine-Tuning. Ref-RL comprises approximately 2K expert-refined samples for Reference-Aligned Group Relative Policy Optimization. Table~\ref{tab:supp-data-statistics} summarizes their sizes and training roles.

\begin{figure*}[!ht]
  \centering
  \includegraphics[
    width=\linewidth,
    height=0.52\textheight,
    keepaspectratio
  ]{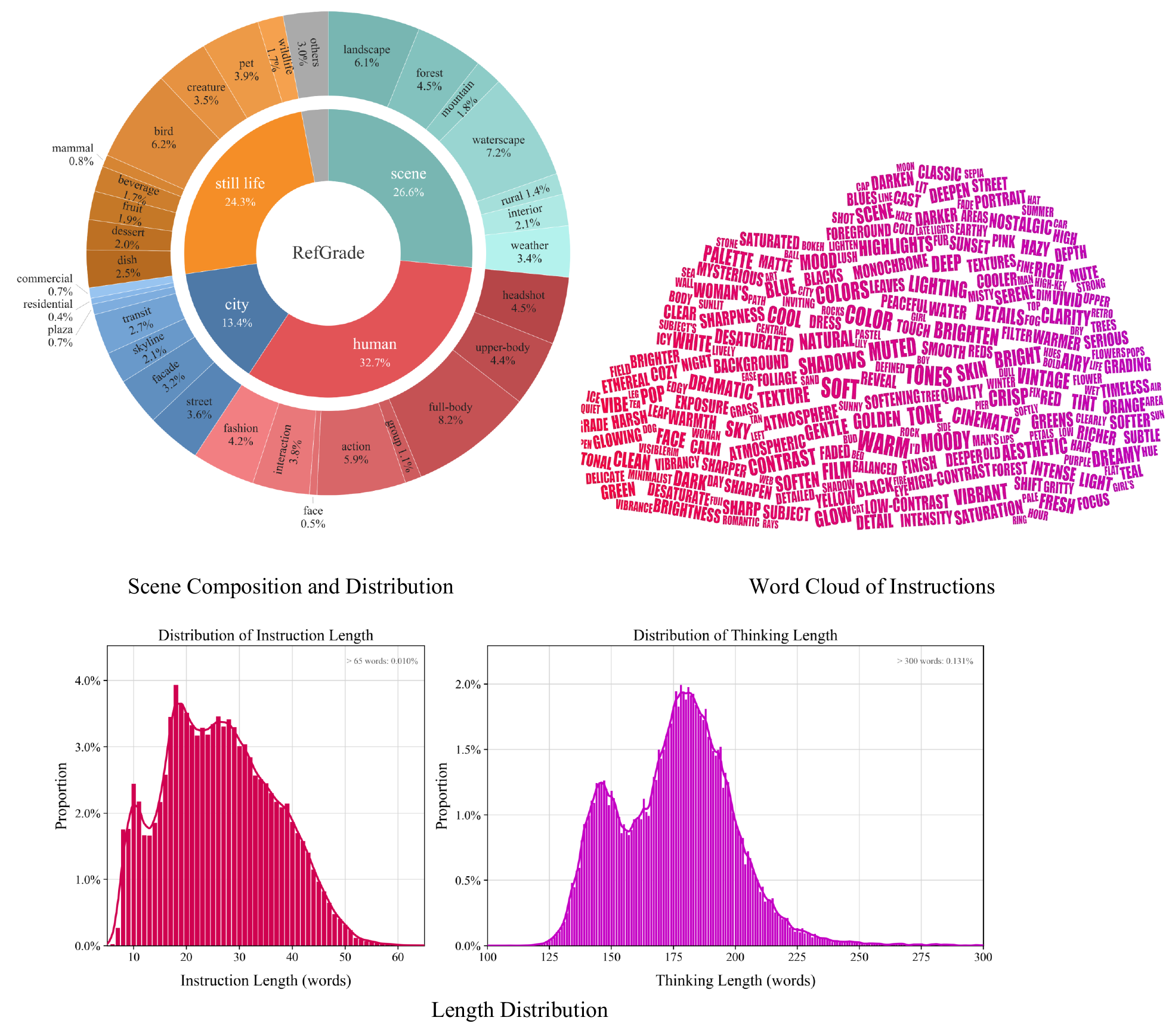}
  \caption{Source-image scene composition, instruction word cloud,
    and instruction and reasoning length distributions.}
  \label{fig:supp-category}
\end{figure*}

The upper-left panel of Figure~\ref{fig:supp-category} illustrates the scene composition of the source-image collection.
The four principal categories are portrait (32.7\%, e.g., headshots, full-body portraits, and action shots), scene (26.6\%, e.g., landscapes, waterscapes, and forests), still life (24.3\%, e.g., food, dishes, and pets), and city (13.4\%, e.g., streets, facades, and transit).
These categories encompass fine-grained subcategories spanning diverse photographic subjects and settings.

The upper-right panel of Figure~\ref{fig:supp-category} summarizes the vocabulary used to express grading intent after removing common stopwords and generic instruction terms.
The remaining vocabulary covers color attributes, lighting conditions, atmosphere, and visual style.
The lower panels present the instruction and chain-of-thought (CoT) length distributions.
Instruction lengths are primarily concentrated between approximately 10 and 45 words, covering common forms of expression ranging from brief requests to more detailed descriptions. CoT lengths are mainly concentrated between approximately 130 and 210 words, with two distinct peaks near 145 and 180 words.

\subsection{Comparison with Existing Datasets}
\label{sec:supp-dataset-comparison}
Table~\ref{tab:supp-dataset-comparison} compares datasets in terms of photographic data, combined reference-image and textual conditioning, reasoning annotations, Lightroom configurations, non-fixed resolution, and continual pre-training support.

\subsection{Additional Data Construction Details}
\label{sec:supp-data-construction}

\noindent\textbf{Annotation Quality Control for Ref-CPT.}
To promote annotation consistency, the four captioning models follow a shared prompting scheme with common output-format and terminology requirements (Figure~\ref{fig:supp-prompt-caption}).
All generated captions undergo validation by a shared VLM-as-Judge verifier conditioned on the source image, editing parameters, and rendered result.
The verifier assesses three aspects: \textbf{1) Format compliance.} Whether captions conform to the prescribed output structure. \textbf{2) Descriptive specificity.} Whether captions characterize parameter-induced appearance changes rather than generic image semantics or aesthetic judgments. \textbf{3) Editing faithfulness.} Whether the described changes are supported by the source and rendered images and consistent with the editing parameters.
The verification prompt is provided in Figure~\ref{fig:supp-prompt-caption-verifier}.
Captions that fail validation are regenerated and re-evaluated for up to three rounds.
Remaining failures undergo manual review, while clearly invalid outputs are discarded.
This process applies consistent quality-control criteria across annotation sources while preserving diversity in wording and expression.

\noindent\textbf{Human-in-the-Loop Aesthetic Screening for Ref-SFT.}
Screening is performed after reference-image construction and before user-instruction and reasoning-trajectory generation.
Each candidate sample is independently reviewed by five volunteers against two criteria:
\textbf{1) Reference alignment.} Whether the constructed reference and target retouched image exhibit consistent key visual attributes, and whether the reference provides transferable cues for the source image, following Table~\ref{tab:supp-inspection-guide}.
\textbf{2) Aesthetic quality.} Whether the rendered results are aesthetically acceptable, without noticeable deviations from human aesthetic preferences.
A sample proceeds to subsequent annotation if at least three reviewers judge it to satisfy both criteria.
Otherwise, the underlying image assets are returned to the data generation and reference construction stage for reuse, and newly constructed candidates undergo the same screening.

\begin{table*}[t]
  \centering
  \caption{Overview of the training resources. Sizes are approximate and exclude dynamically derived samples.}
  \label{tab:supp-data-statistics}

  \setlength{\tabcolsep}{5pt}
  \renewcommand{\arraystretch}{1.0}

  \begin{tabularx}{\textwidth}{
    @{}lcl>{\raggedright\arraybackslash}X@{}
  }
    \toprule
    Dataset & Samples & Training stage & Principal content \\
    \midrule
    Ref-CPT & 200K & PG-CPT
    & Parameter-grounded state-change captions \\
    Ref-SFT & 50K & RG-SFT
    & Reference-grounded instructions, reasoning, and parameters \\
    Ref-RL & 2K & RA-GRPO
    & Expert-refined cases for rendered-outcome optimization \\
    \bottomrule
  \end{tabularx}
\end{table*}

\begin{table*}[t]
  \centering
  \caption{Comparison with existing datasets. Res. denotes non-fixed resolution and CPT denotes continual pre-training support.}
  \label{tab:supp-dataset-comparison}

  \setlength{\tabcolsep}{3pt}
  \begin{tabular}{@{}ccccccc@{}}
    \toprule
    Dataset & Real photos & Reference and text & CoT labels
    & Lightroom config. & Res. & CPT \\
    \midrule
    InstructPix2Pix~\cite{brooks2023instructpix2pix}
    & No & No & No & No & No & No \\
    MagicBrush~\cite{zhang2023magicbrush}
    & Yes & No & No & No & No & No \\
    MGIE~\cite{fu2023guiding}
    & Yes & No & No & No & No & No \\
    ImgEdit~\cite{ye2025imgedit}
    & Yes & No & No & No & Yes & No \\
    MIT-Adobe FiveK~\cite{bychkovsky2011learning}
    & Yes & No & No & Yes & Yes & No \\
    PPR10K~\cite{liang2021ppr10k}
    & Yes & No & No & Yes & Yes & No \\
    MMArt~\cite{lin2025jarvisart}
    & Yes & No & Yes & Yes & Yes & No \\
    ArtEdit~\cite{lin2025jarvisevo}
    & Yes & No & Yes & Yes & Yes & No \\
    Ref datasets (ours)
    & Yes & Yes & Yes & Yes & Yes & Yes \\
    \bottomrule
  \end{tabular}
\end{table*}

\noindent\textbf{Expert Refinement and Consensus Validation for Ref-RL.}
We curate a reference library from high-quality photographs purchased with licenses covering the intended uses.
A free color-grading service collects real user requests, allowing users to select a reference and submit a source image with natural-language instructions.
Submissions are used only after disclosure of the research purpose and explicit user authorization.
Experts perform parameterized editing according to each request, retaining configurations that reproduce the corresponding target images.

Each candidate is independently assessed by 20 reviewers spanning different age groups and levels of imaging expertise, including general users, photography enthusiasts, and imaging professionals.
Editors do not review their own outputs.
Reviewers examine the source image, reference image, user instruction, and edited result against two criteria:
\textbf{1) Reference and instruction alignment.} Whether the result draws on the reference's overall style and tone while fulfilling the user's attribute-specific preferences and requested adjustments.
\textbf{2) Aesthetic quality.} Whether the result is visually coherent and free of conspicuous defects.
Both criteria receive binary pass/fail judgments, and acceptance requires unanimous approval on both.
Consensus denotes panel agreement on task fulfillment and visual acceptability, rather than a uniquely optimal edit.

Failed candidates are returned to the editor with specific feedback and revised without altering the user request.
The same panel independently reassesses revisions, and only unanimously approved samples are retained.
Ref-RL comprises approximately 2K consensus-validated, expert-refined samples, disjoint from Ref-Bench.
Samples dynamically derived during training are counted separately.

\subsection{Representative Data Samples}
\label{sec:supp-data-samples}

\begin{figure*}[p]
    \centering
    \includegraphics[
        width=\textwidth,
        height=0.85\textheight,
        keepaspectratio
    ]{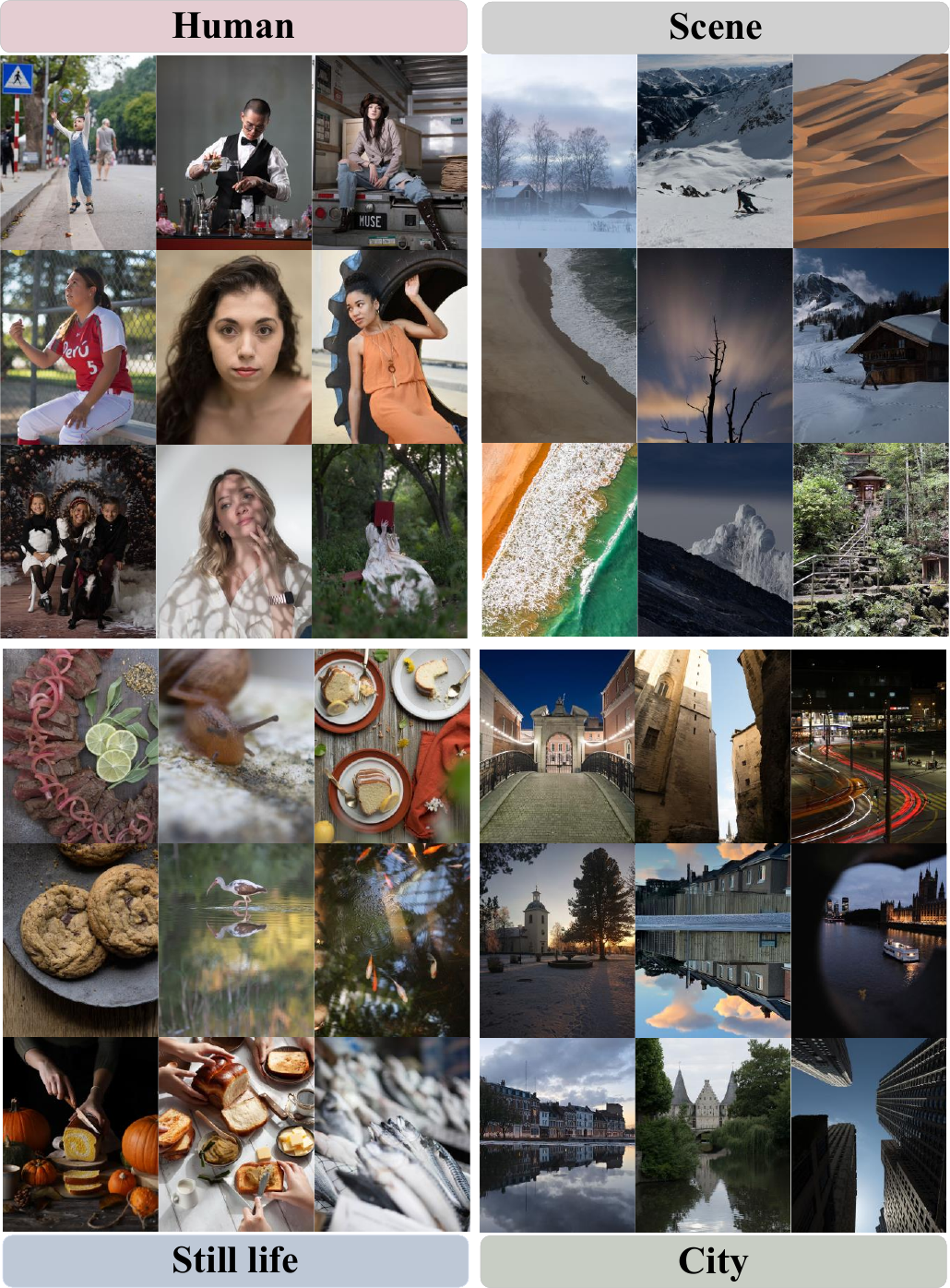}
    \caption{Representative photographic scenes in the Ref datasets.}
    \label{fig:supp-4scene}
\end{figure*}

\begin{figure*}[p]
    \centering
    \includegraphics[
        width=\textwidth,
        height=0.85\textheight,
        keepaspectratio
    ]{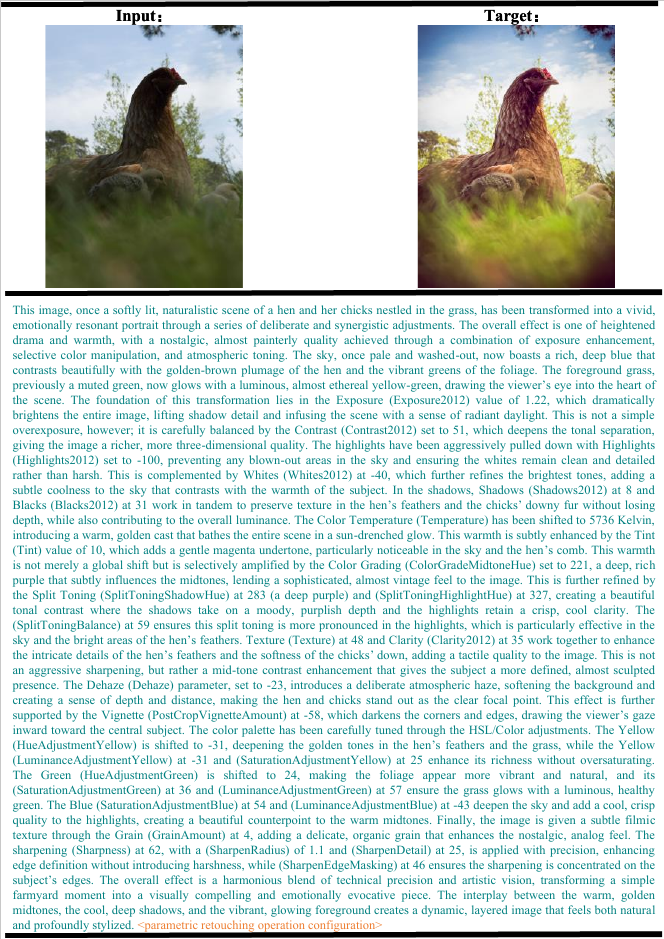}
    \caption{Ref-CPT examples with source images, rendered outcomes, editing configurations, and state-change captions.}
    \label{fig:supp-cptData}
\end{figure*}

\begin{figure*}[p]
    \centering
    \includegraphics[
        width=\textwidth,
        height=0.85\textheight,
        keepaspectratio
    ]{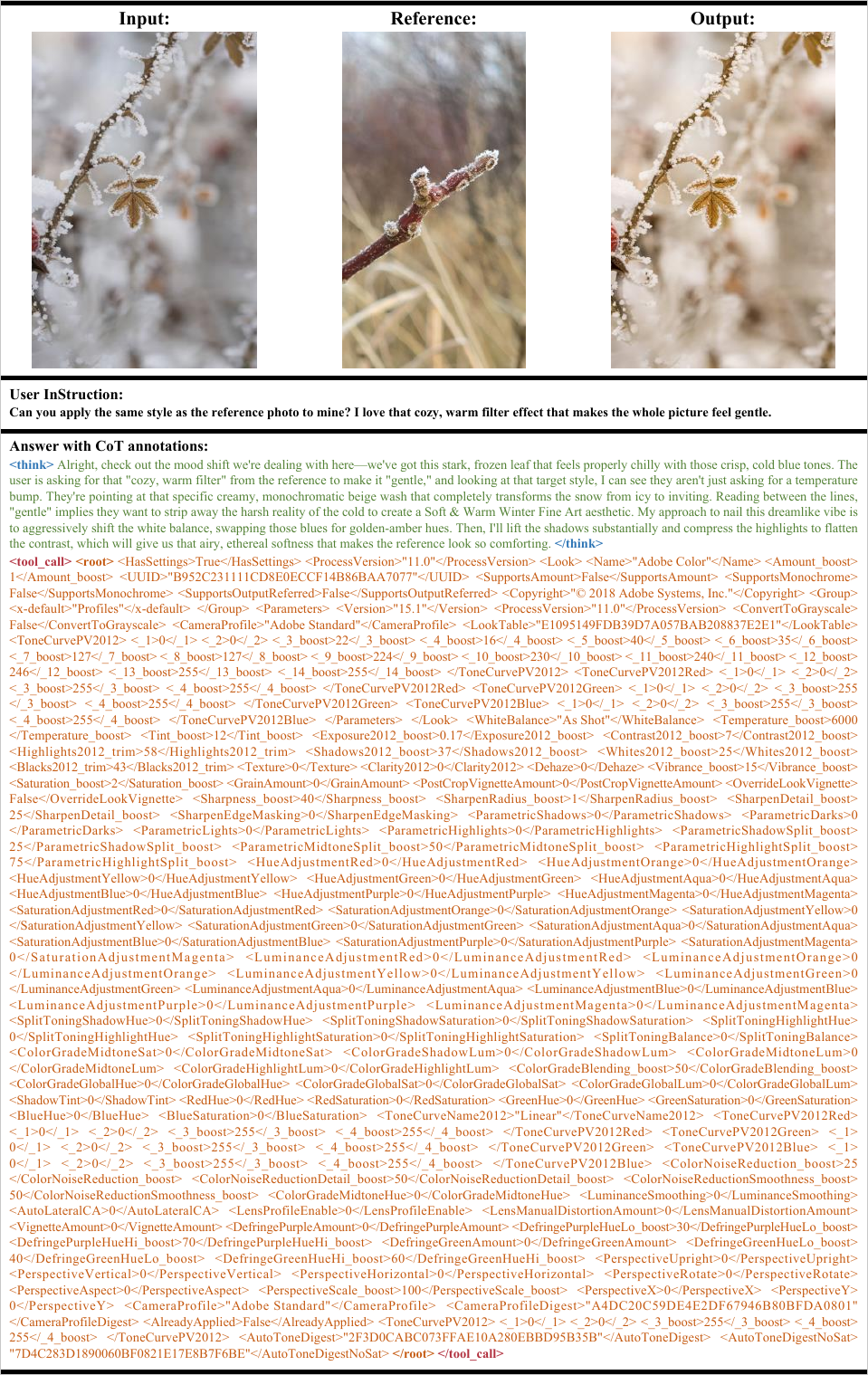}
    \caption{Ref-SFT examples with source and reference images, instructions, reasoning trajectories, parameter configurations, and rendered outcomes.}
    \label{fig:supp-sftData}
\end{figure*}

\begin{figure*}[p]
    \centering
    \includegraphics[
        width=\textwidth,
        height=0.85\textheight,
        keepaspectratio
    ]{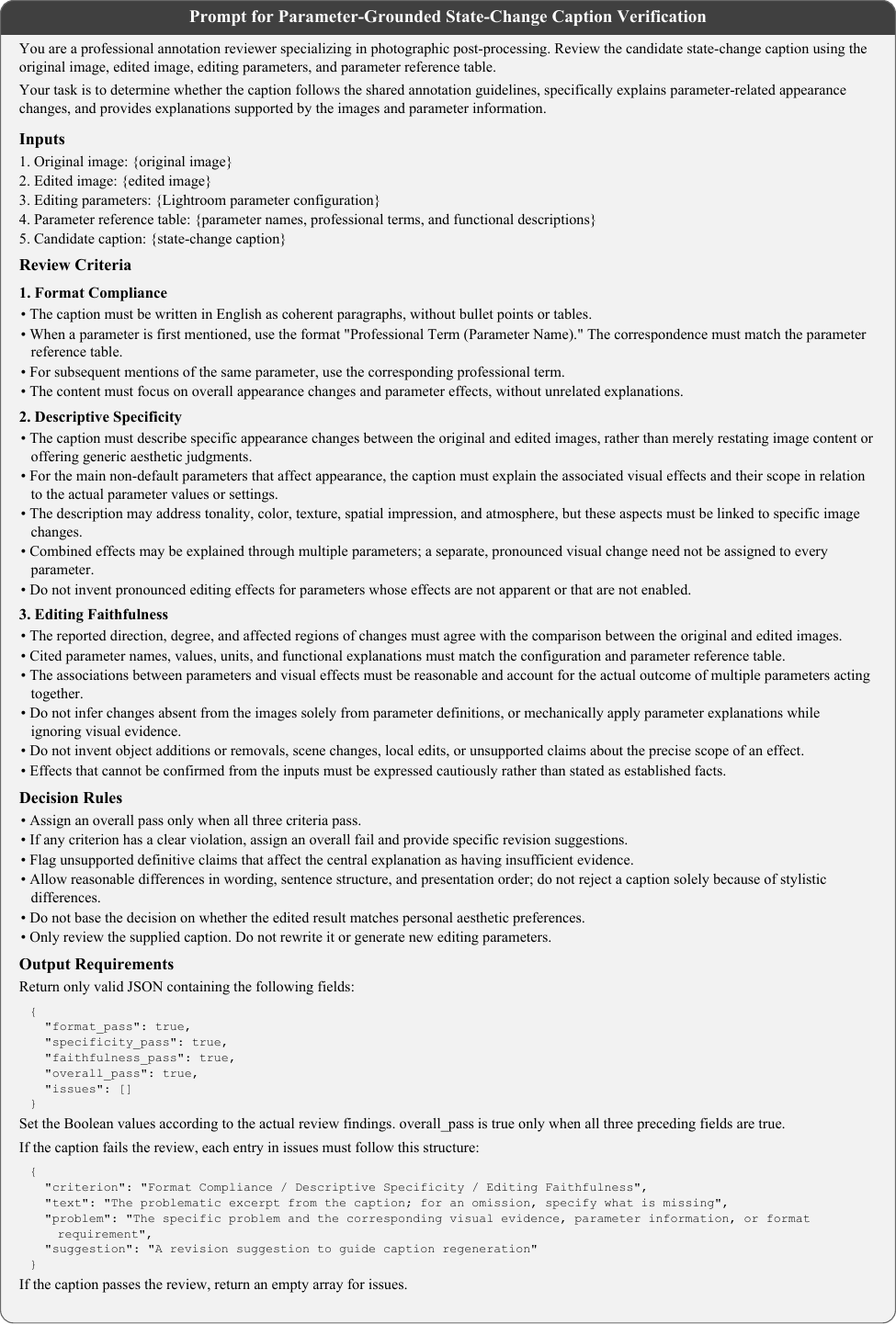}
    \caption{Prompt template for verifying state-change caption validity and faithfulness.}
    \label{fig:supp-prompt-caption-verifier}
\end{figure*}

Figures~\ref{fig:supp-4scene} and~\ref{fig:supp-cptData} illustrate scene coverage and Ref-CPT sample structure.
Figures~\ref{fig:supp-sftData} present Ref-SFT examples.

\subsection{Prompt Templates for Data Construction}
\label{sec:supp-data-prompts}

\begin{figure*}[p]
    \centering
    \includegraphics[
        width=\textwidth,
        height=0.85\textheight,
        keepaspectratio
    ]{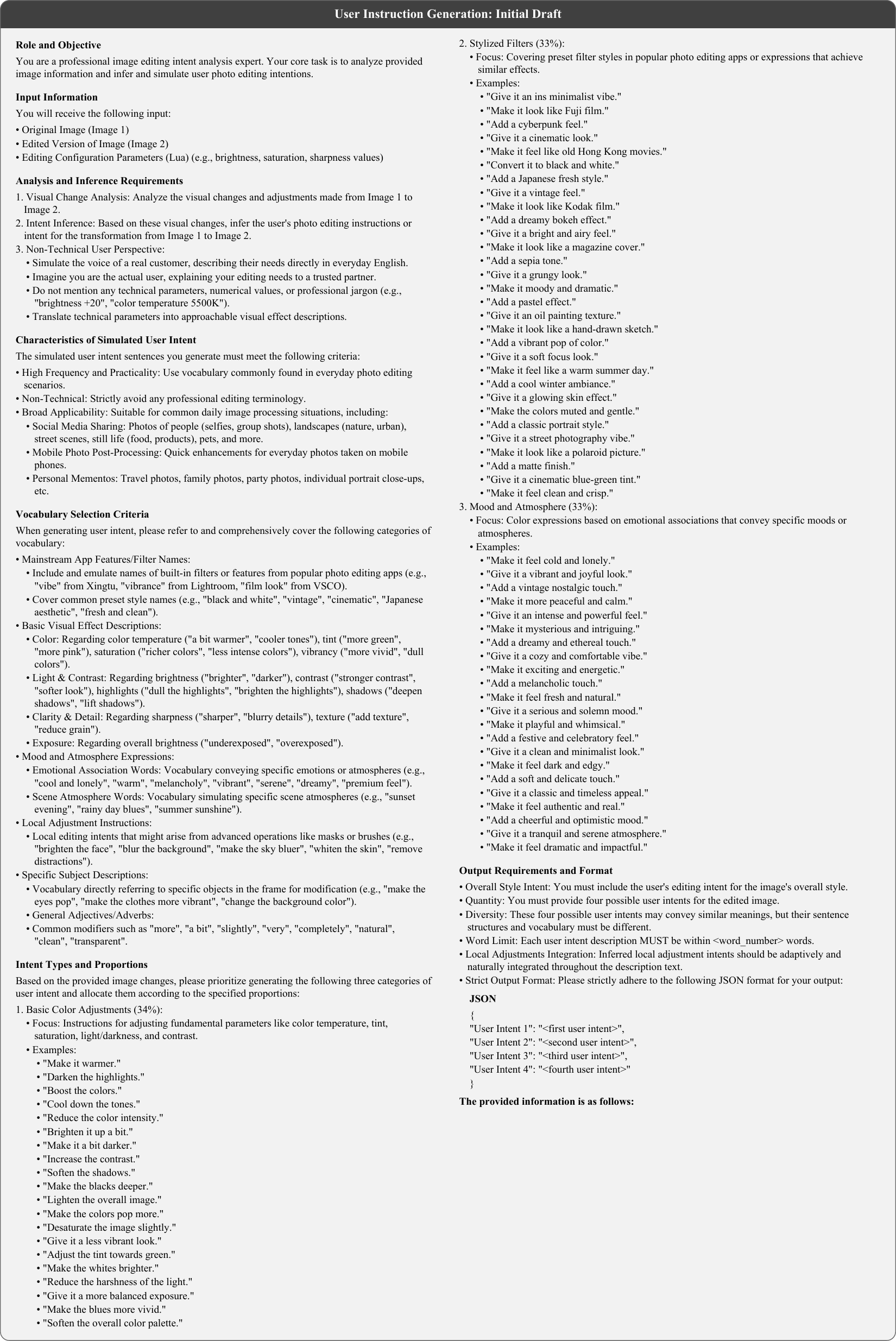}
    \caption{Prompt template for initial user-instruction generation.}
    \label{fig:supp-prompt-instruction}
\end{figure*}

\begin{figure*}[p]
    \centering
    \includegraphics[
        width=\textwidth,
        height=0.85\textheight,
        keepaspectratio
    ]{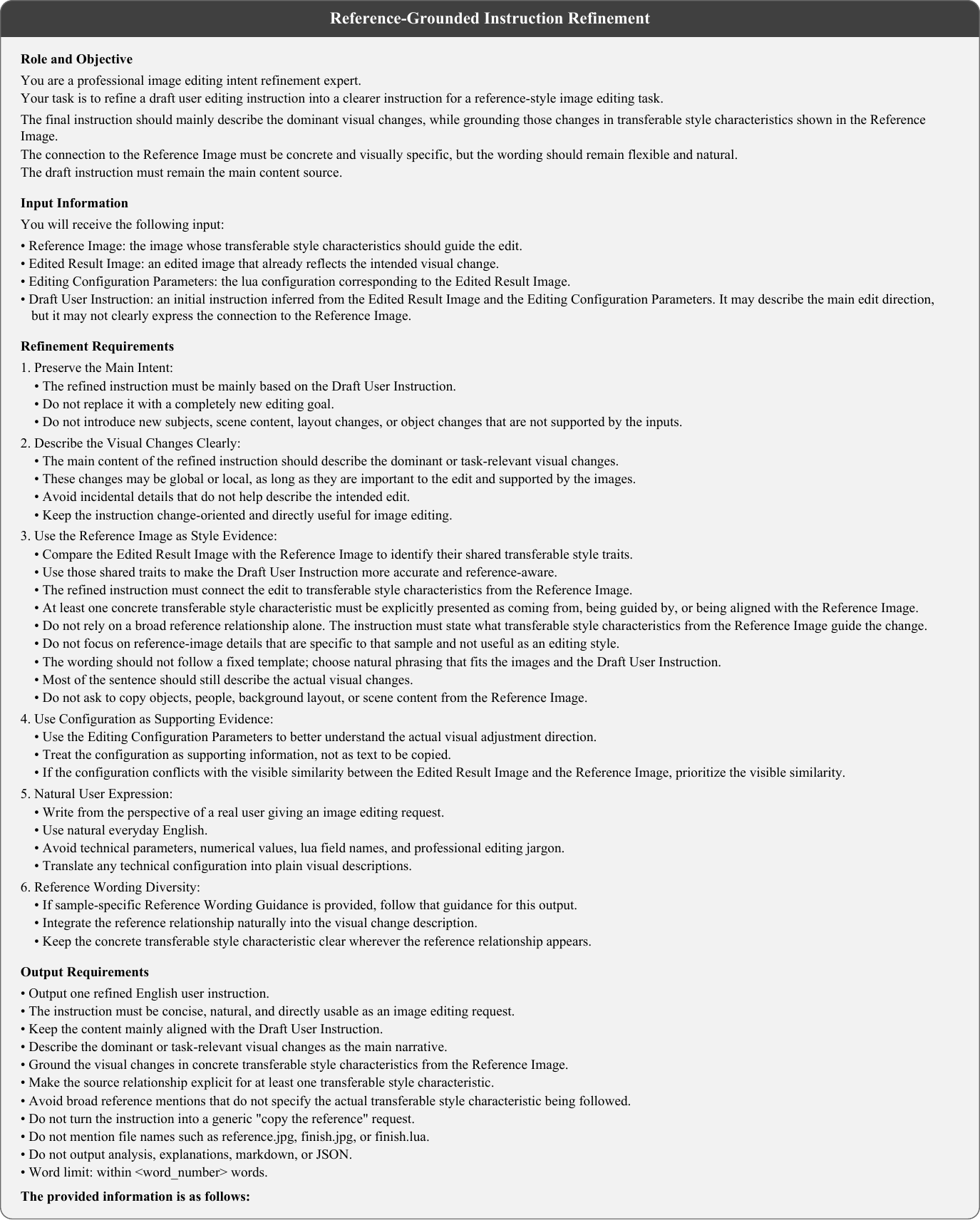}
    \caption{Prompt template for reference-grounded user-instruction refinement.}
    \label{fig:supp-prompt-instruction-refinement}
\end{figure*}

\begin{figure*}[p]
    \centering
    \includegraphics[
        width=\textwidth,
        height=0.85\textheight,
        keepaspectratio
    ]{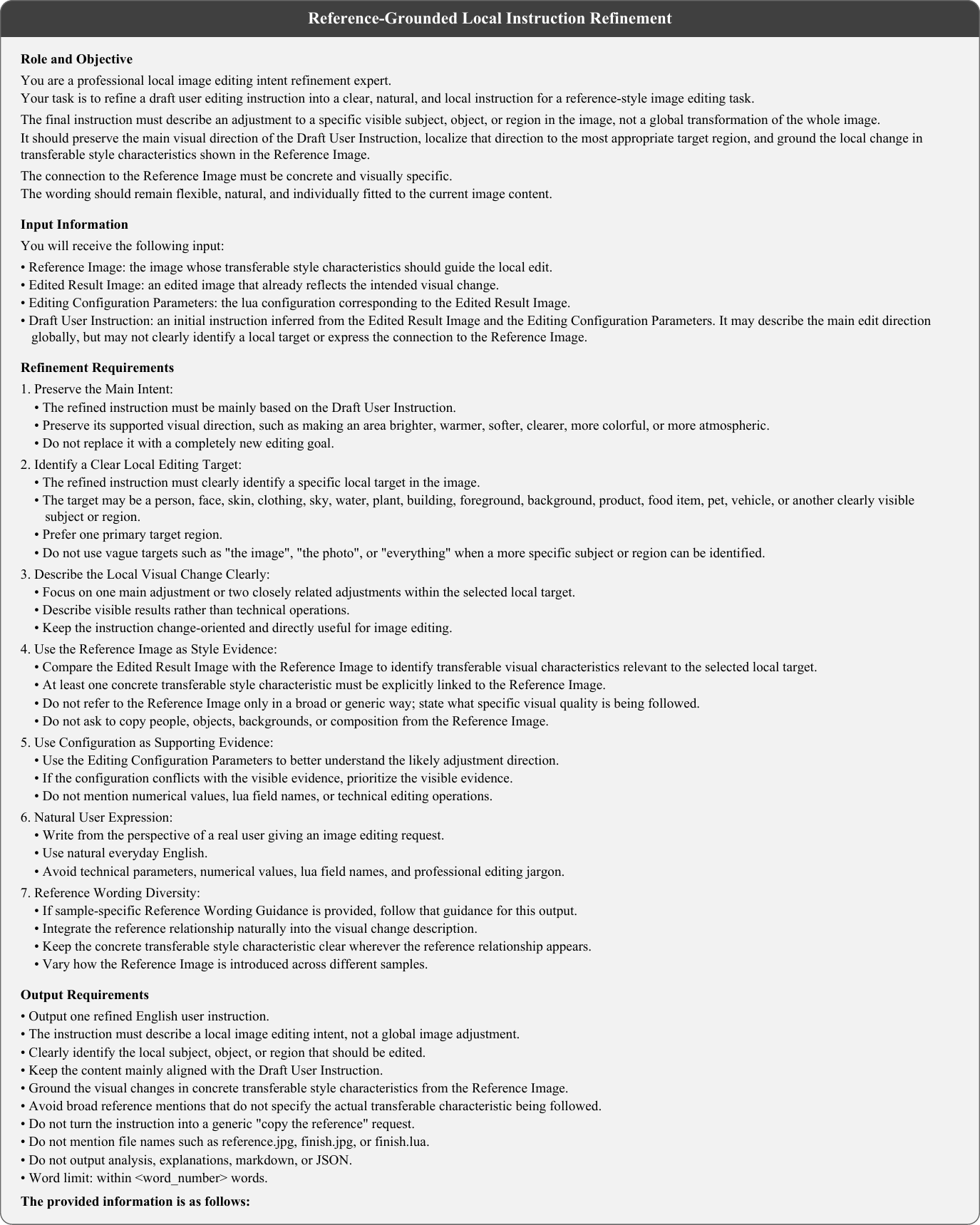}
    \caption{Prompt template for reference-grounded local user-instruction refinement.}
    \label{fig:supp-prompt-instruction-local}
\end{figure*}

\begin{figure*}[p]
    \centering
    \includegraphics[
        width=\textwidth,
        height=0.85\textheight,
        keepaspectratio
    ]{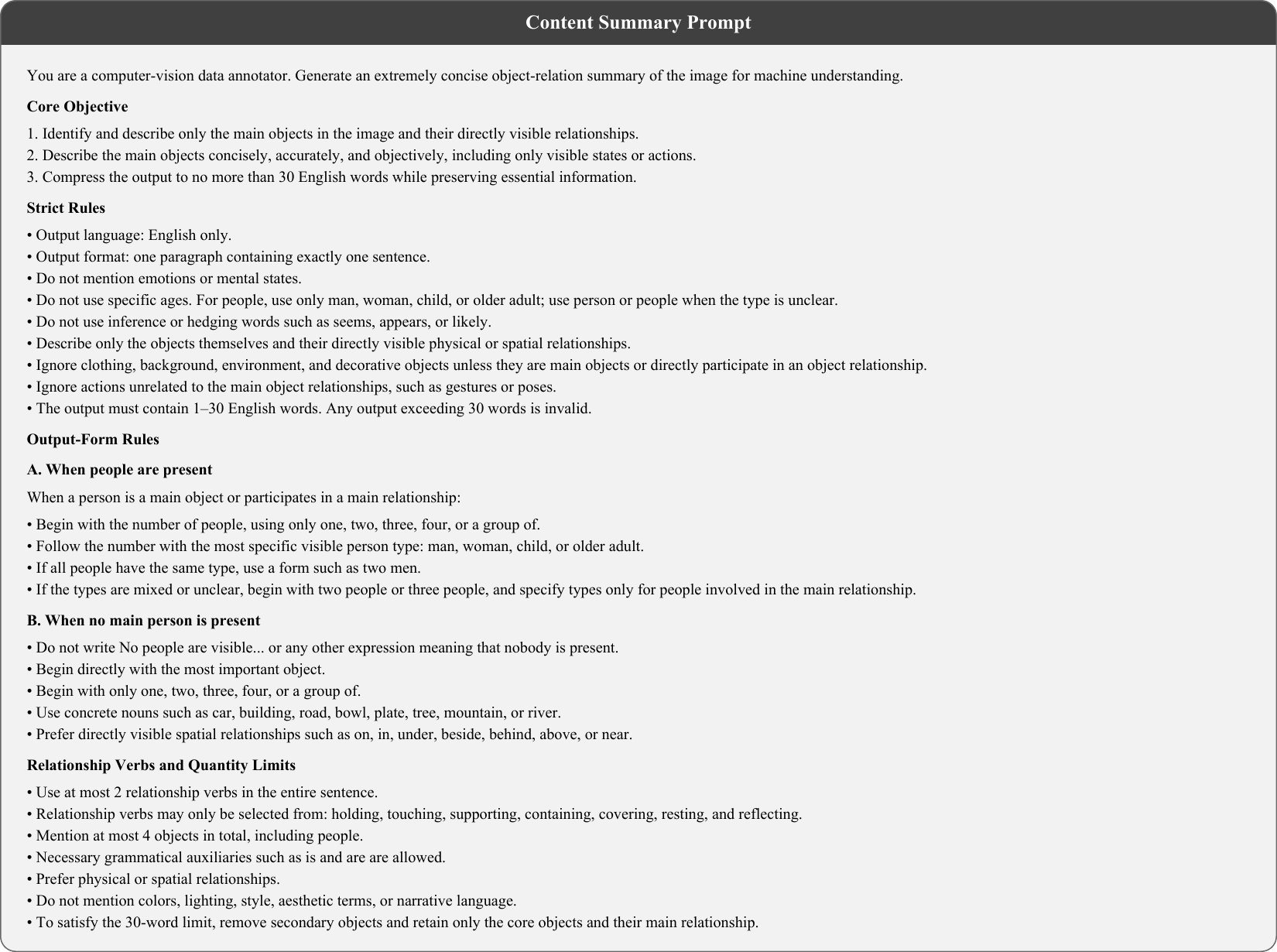}
    \caption{Prompt template for content summarization in semantic retrieval.}
    \label{fig:supp-prompt-content}
\end{figure*}

\begin{figure*}[p]
    \centering
    \includegraphics[
        width=\textwidth,
        height=0.85\textheight,
        keepaspectratio
    ]{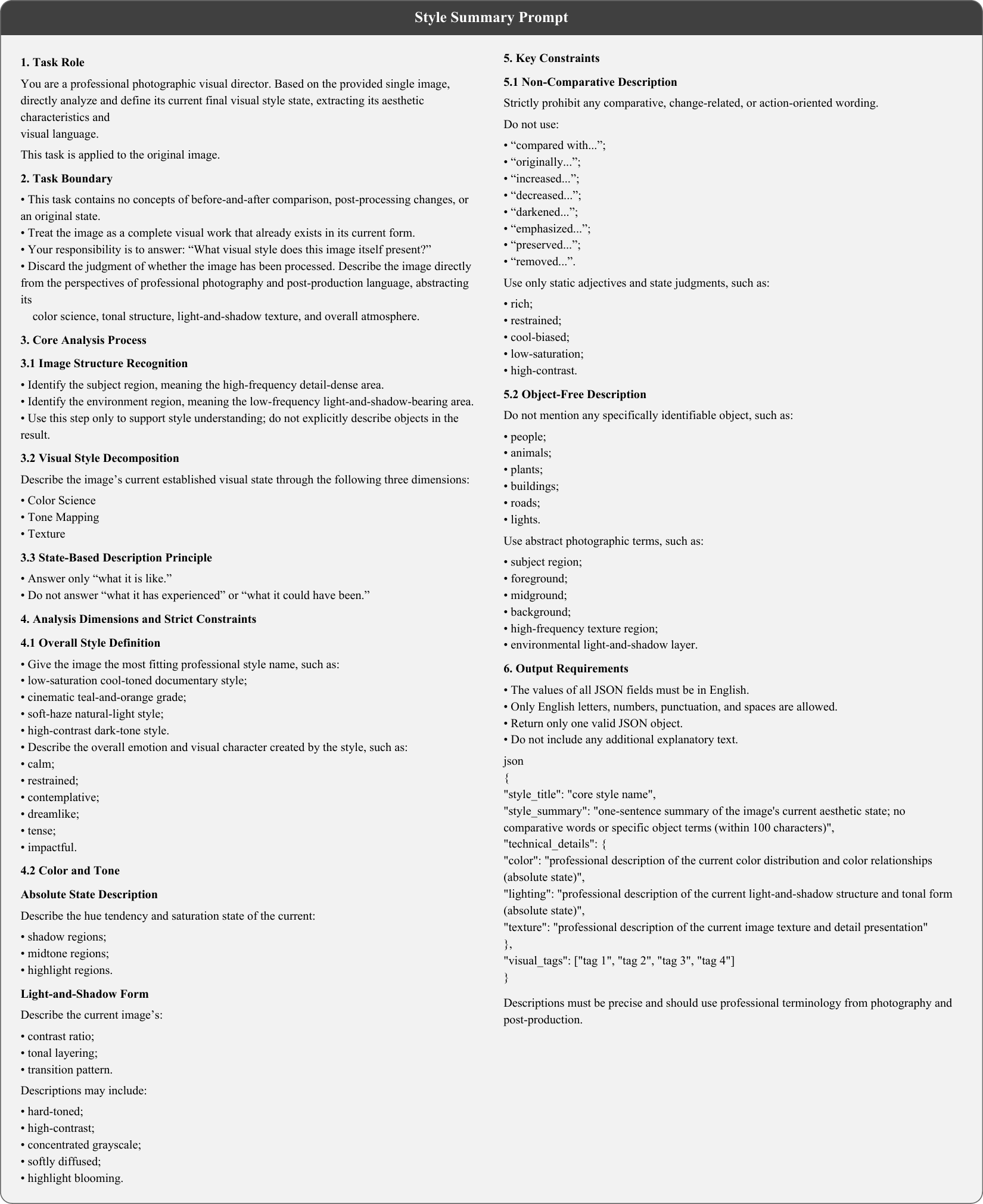}
    \caption{Prompt template for grading-style summarization in reference reranking.}
    \label{fig:supp-prompt-style}
\end{figure*}

\begin{figure*}[p]
  \centering
  \includegraphics[
    width=\textwidth,
    height=0.9\textheight,
    keepaspectratio
  ]{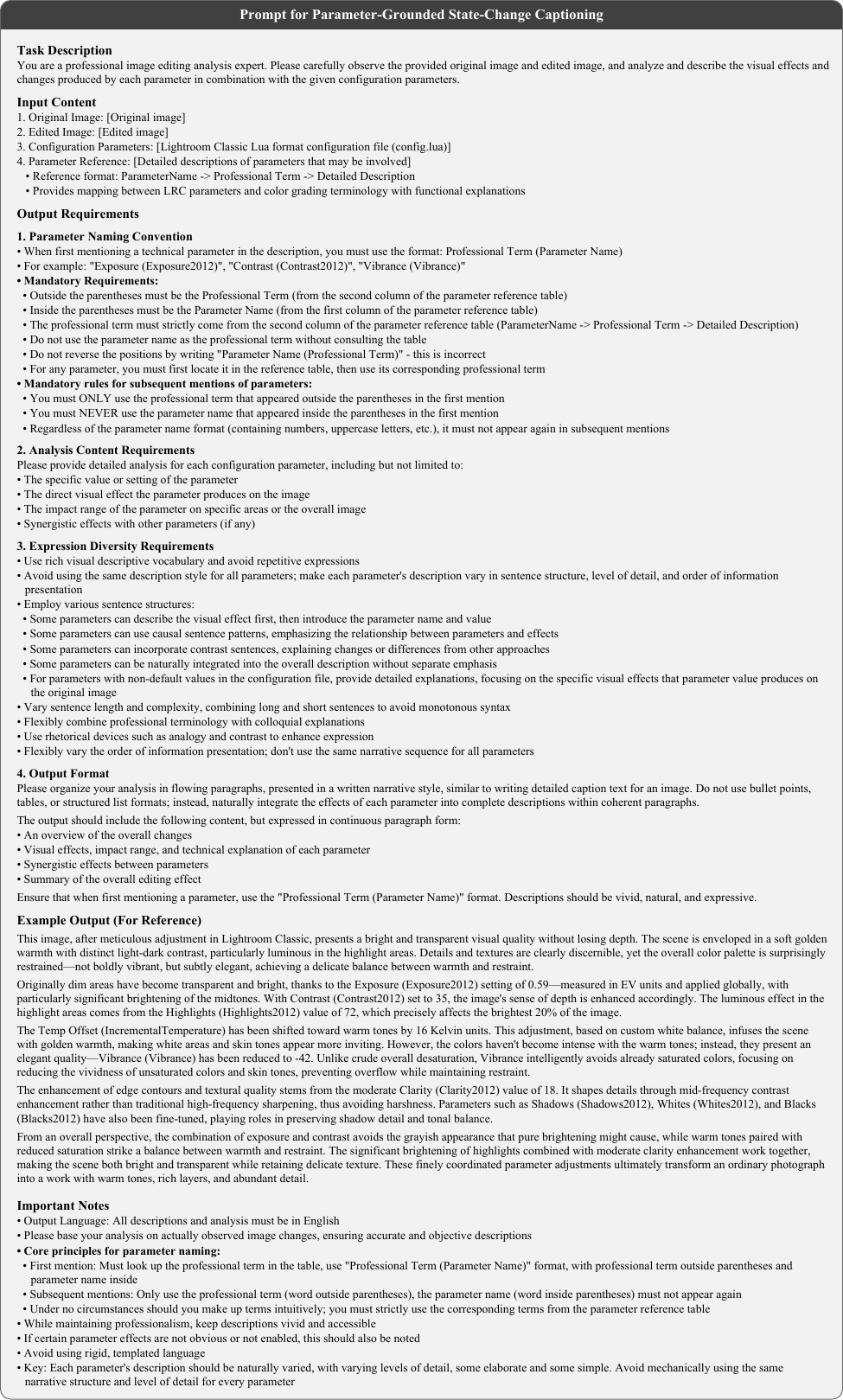}
  \caption{Prompt template for parameter-grounded state-change captioning.}
  \label{fig:supp-prompt-caption}
\end{figure*}

\begin{figure*}[p]
    \centering
    \includegraphics[
        width=\textwidth,
        height=0.85\textheight,
        keepaspectratio
    ]{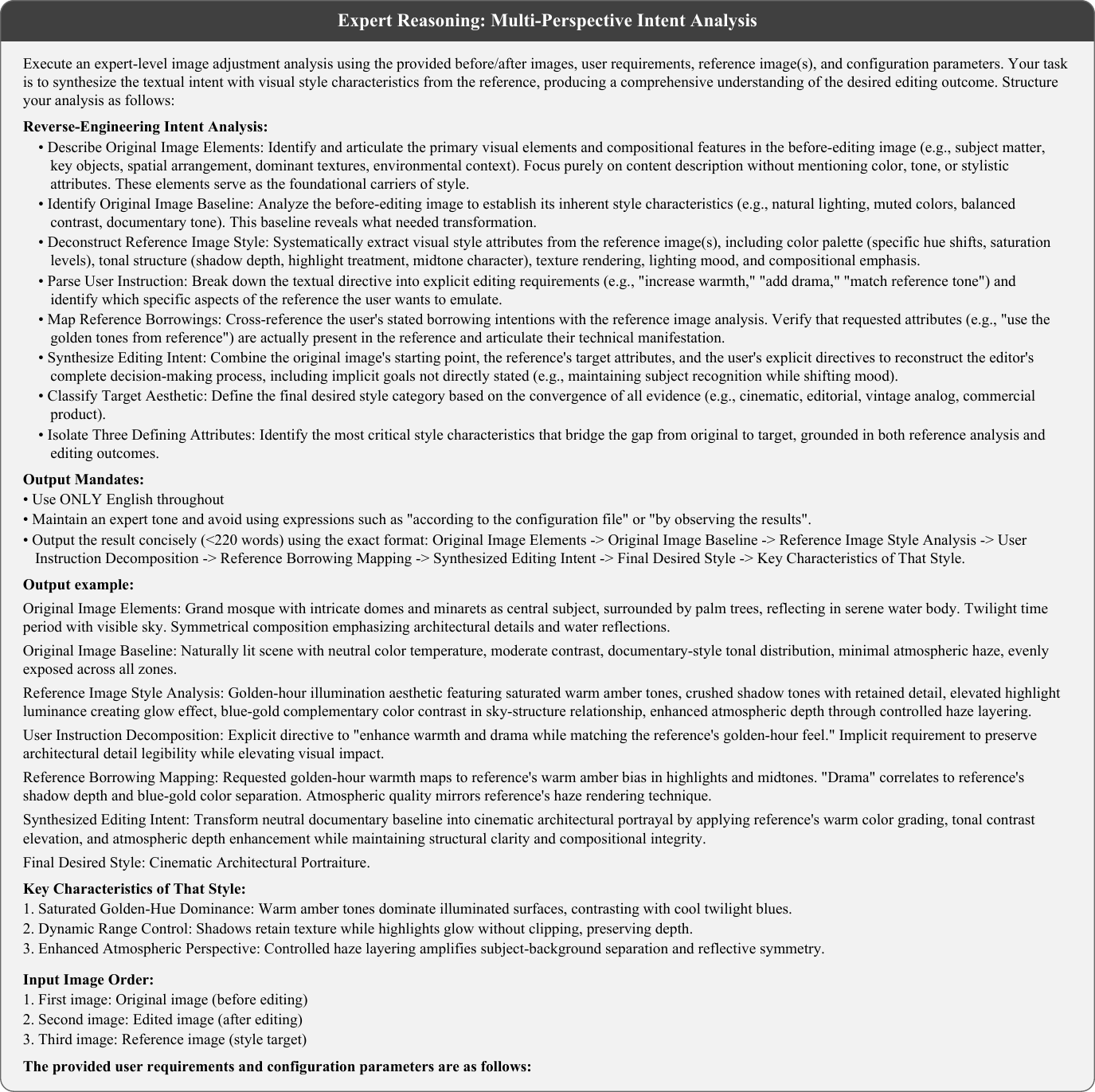}
    \caption{Prompt template for structured interpretation of grading intent.}
    \label{fig:supp-prompt-intent}
\end{figure*}

\begin{figure*}[p]
    \centering
    \includegraphics[
        width=\textwidth,
        height=0.85\textheight,
        keepaspectratio
    ]{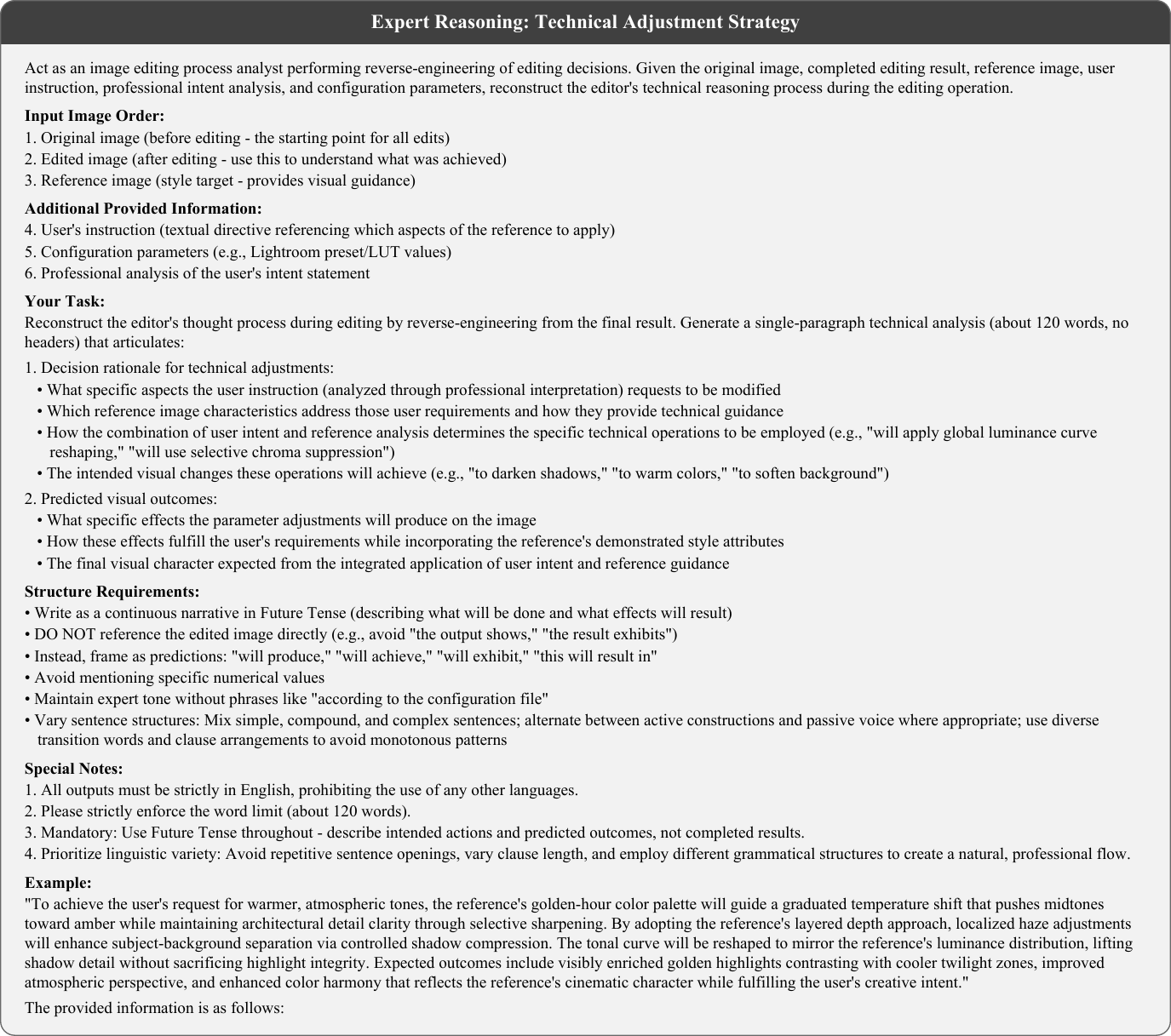}
    \caption{Prompt template for mapping grading intent to parameter decisions.}
    \label{fig:supp-prompt-parameter}
\end{figure*}

\begin{figure*}[p]
    \centering
    \includegraphics[
        width=\textwidth,
        height=0.85\textheight,
        keepaspectratio
    ]{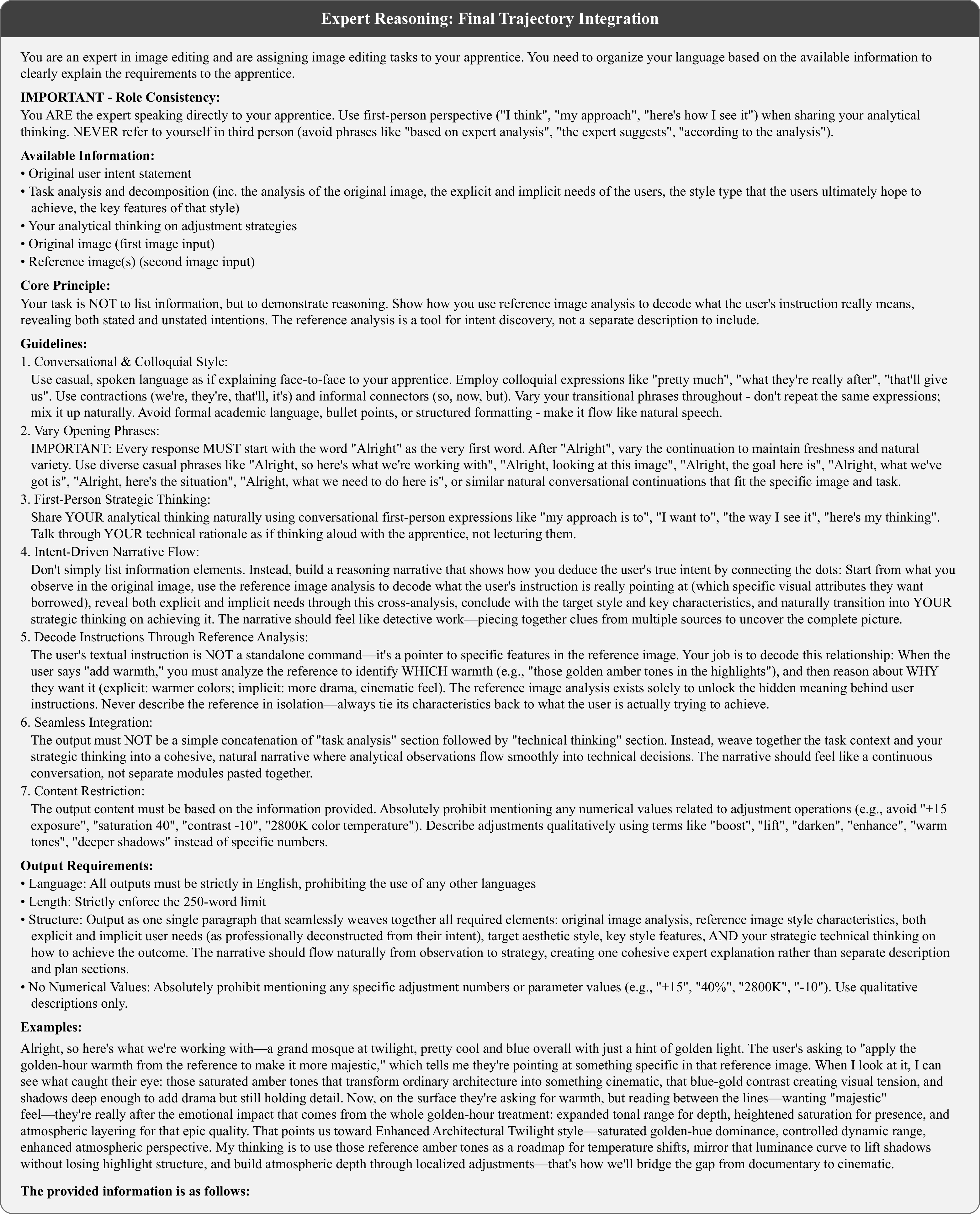}
    \caption{Prompt template for integrating intent analysis and parameter decisions into a reasoning trajectory.}
    \label{fig:supp-prompt-reasoning}
\end{figure*}

Figure~\ref{fig:supp-prompt-caption} presents the state-change captioning prompt.
Figures~\ref{fig:supp-prompt-content} and~\ref{fig:supp-prompt-style} provide the summary prompts used in reference retrieval.
Figures~\ref{fig:supp-prompt-instruction}, \ref{fig:supp-prompt-instruction-refinement}, and~\ref{fig:supp-prompt-instruction-local} present the prompt templates for initial user-instruction generation, reference-grounded instruction refinement, and reference-grounded local instruction refinement, respectively.
Figures~\ref{fig:supp-prompt-intent}, \ref{fig:supp-prompt-parameter}, and~\ref{fig:supp-prompt-reasoning} correspond to intent analysis, parameter-decision mapping, and reasoning-trajectory integration.
Quality-verification, reward, derivation, and evaluation prompts are referenced separately in their respective methodological sections.

\section{Methodological Details}
\label{sec:supp-methodology}

\subsection{Task-Alignment Reward}
\label{sec:supp-task-reward}

The task-alignment reward combines instruction following and reference matching.

\noindent\textbf{1) Instruction Following.}
A vision-language evaluator assesses instruction fulfillment using the source image, reference image, rendered output, user instruction, and executed parameter configuration.
The reference provides stylistic and tonal context, while the instruction determines the attributes and adjustments to evaluate.
Parameter settings provide auxiliary evidence of adjustment direction and scope, but fulfillment depends on visible outcomes rather than parameter values alone.

Using the prompt in Figure~\ref{fig:supp-prompt-task-reward}, the evaluator directly scores four dimensions: color, light and shadow, technical adjustments, and high-level style.
Each applicable dimension receives a score of 100, 70, 35, or 0, corresponding to full fulfillment, fulfillment with minor discrepancies, partial fulfillment, or failure, respectively.
Dimensions without corresponding requirements are marked inapplicable, whereas those lacking sufficient evidence are marked unassessable.
Both are excluded from aggregation, with the reason recorded.
The instruction-following component $\mathcal{V}_{\mathrm{inst}}$ is the equal-weight mean of the remaining scores divided by 100, yielding a value in $[0,1]$.
If no dimension is assessable, the evaluation is flagged as invalid and assigned zero.
Quality attributes not specified in the instruction are assessed separately by the visual-quality reward.

\noindent\textbf{2) Reference Matching.}
Reference matching compares each rendered output with the expert-refined target.
Before color conversion, images whose longest side exceeds 512 pixels are proportionally downsampled using area interpolation; smaller images retain their original resolution.

\textit{Color-space conversion.}
Each sRGB channel $C\in[0,1]$ is first linearized:
\begin{equation}
C_{\mathrm{lin}}=
\begin{cases}
C/12.92, & C\leq 0.04045,\\
\left((C+0.055)/1.055\right)^{2.4}, & C>0.04045.
\end{cases}
\end{equation}
The linear RGB values are converted to CIE XYZ under the D65 illuminant:
\begin{equation}
\begin{bmatrix}
X\\Y\\Z
\end{bmatrix}
=
\begin{bmatrix}
0.412453 & 0.357580 & 0.180423\\
0.212671 & 0.715160 & 0.072169\\
0.019334 & 0.119193 & 0.950227
\end{bmatrix}
\begin{bmatrix}
R_{\mathrm{lin}}\\
G_{\mathrm{lin}}\\
B_{\mathrm{lin}}
\end{bmatrix}.
\end{equation}
CIELAB coordinates are then obtained as
\begin{equation}
\begin{aligned}
L^\ast &= 116f(Y/Y_n)-16,\\
a^\ast &= 500\left[f(X/X_n)-f(Y/Y_n)\right],\\
b^\ast &= 200\left[f(Y/Y_n)-f(Z/Z_n)\right],
\end{aligned}
\end{equation}
where $(X_n,Y_n,Z_n)=(0.950456,1,1.088754)$ and
\begin{equation}
f(t)=
\begin{cases}
t^{1/3}, & t>\delta^3,\\
t/(3\delta^2)+4/29, & t\leq\delta^3,
\end{cases}
\qquad \delta=6/29.
\end{equation}
The implementation uses OpenCV's Lab conversion, with the chromatic channels offset by 128 in the 8-bit representation.

\textit{Histogram distance.}
Only the $a$ and $b$ channels are used for matching.
Each channel is discretized into 24 bins, producing a $24\times24$ joint histogram normalized to unit mass.
Bins with probability mass greater than $10^{-8}$ are retained as weighted two-dimensional bin centers.
For candidate $i$, the matching distance is
\begin{equation}
d_i=
\operatorname{EMD}
\left(
H(I_i^{\mathrm{out}}),
H(I_{\mathrm{tgt}})
\right),
\end{equation}
where $H(\cdot)$ denotes the normalized chromatic histogram.
The Earth Mover's Distance is computed using Euclidean distances between bin centers as the transport cost.
Neither the $L$ channel nor pixel coordinates enter this calculation.

\textit{Within-group scoring.}
For each input, the eight rollout candidates are ranked in ascending order of $d_i$.
Equal distances are resolved by preserving the original rollout order.
A measurable candidate with rank $r_i\in\{1,\ldots,8\}$ receives
\begin{equation}
R_i^{\mathrm{match}}=\frac{9-r_i}{8}.
\end{equation}
Thus, scores decrease from $1$ to $0.125$ in increments of $0.125$.
Candidates without a measurable distance because of invalid model outputs receive the minimum score of $0.125$.
Infrastructure failures are excluded from optimization rather than treated as low-ranked candidates.

Finally, the reference-matching score and the instruction-following score are added:
\begin{equation}
R_i^{\mathrm{task}}
=
R_i^{\mathrm{match}}
+
\mathcal{V}_{\mathrm{inst}}
\left(
I_{\mathrm{src}},I_{\mathrm{ref}},
I_i^{\mathrm{out}},Q,P^{(i)}
\right).
\end{equation}

\subsection{Visual-Quality Reward}
\label{sec:supp-quality-reward}

The visual-quality reward measures edit-induced damage independently of task alignment. Most of the $K=14$ criteria assess preservation properties, and a criterion is considered violated only when the edit visibly degrades the corresponding property relative to the source image. The source image therefore serves as the baseline for identifying editing damage, while global tonal or stylistic changes are not penalized unless they cause a concrete criterion-specific defect.

\begin{figure*}[p]
    \centering
    \includegraphics[
        width=\textwidth,
        height=0.85\textheight,
        keepaspectratio
    ]{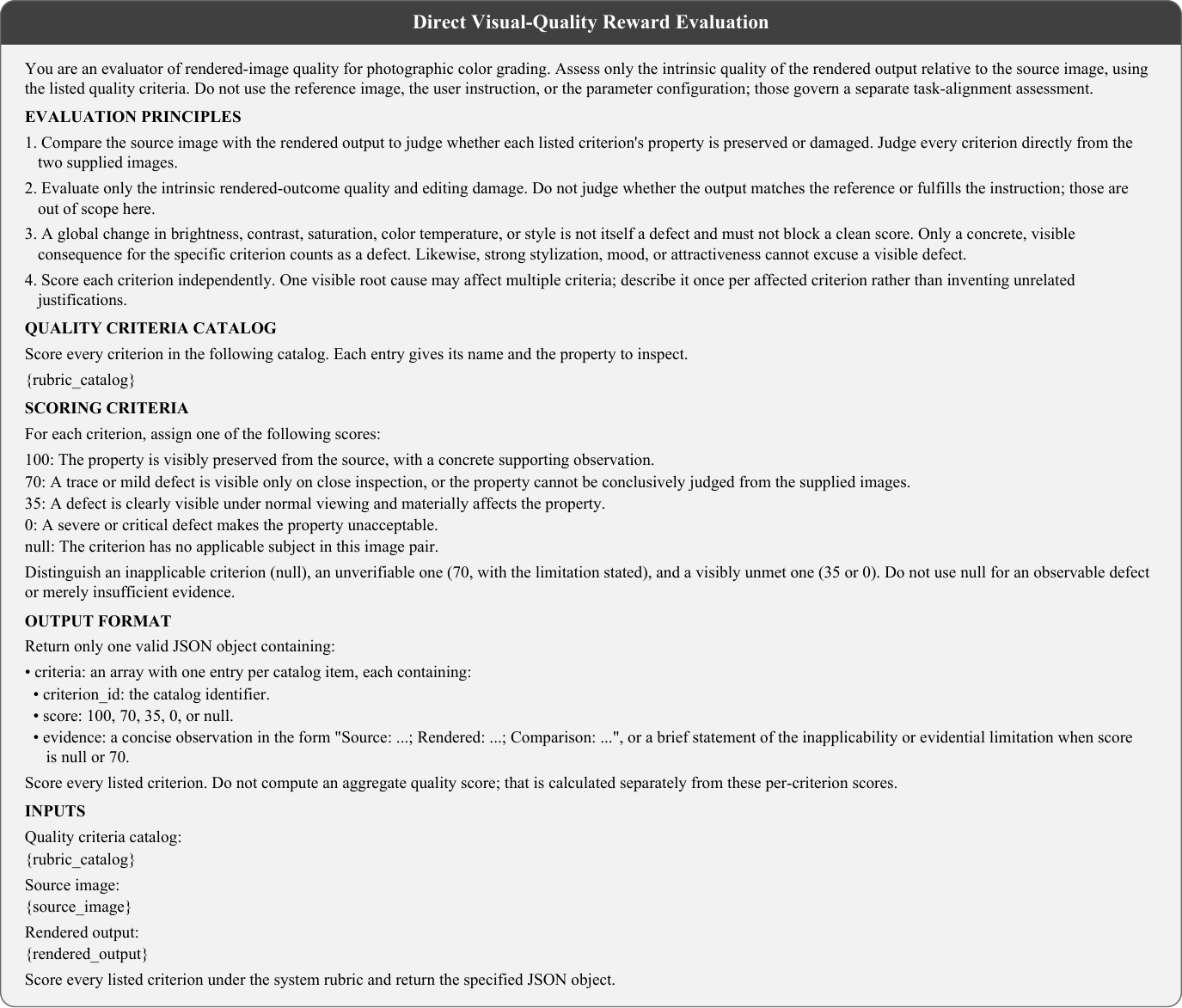}
    \caption{Prompt template for the fourteen-criterion visual-quality reward.}
    \label{fig:supp-prompt-quality-reward}
\end{figure*}

\begin{figure*}[p]
    \centering
    \includegraphics[
        width=\textwidth,
        height=0.85\textheight,
        keepaspectratio
    ]{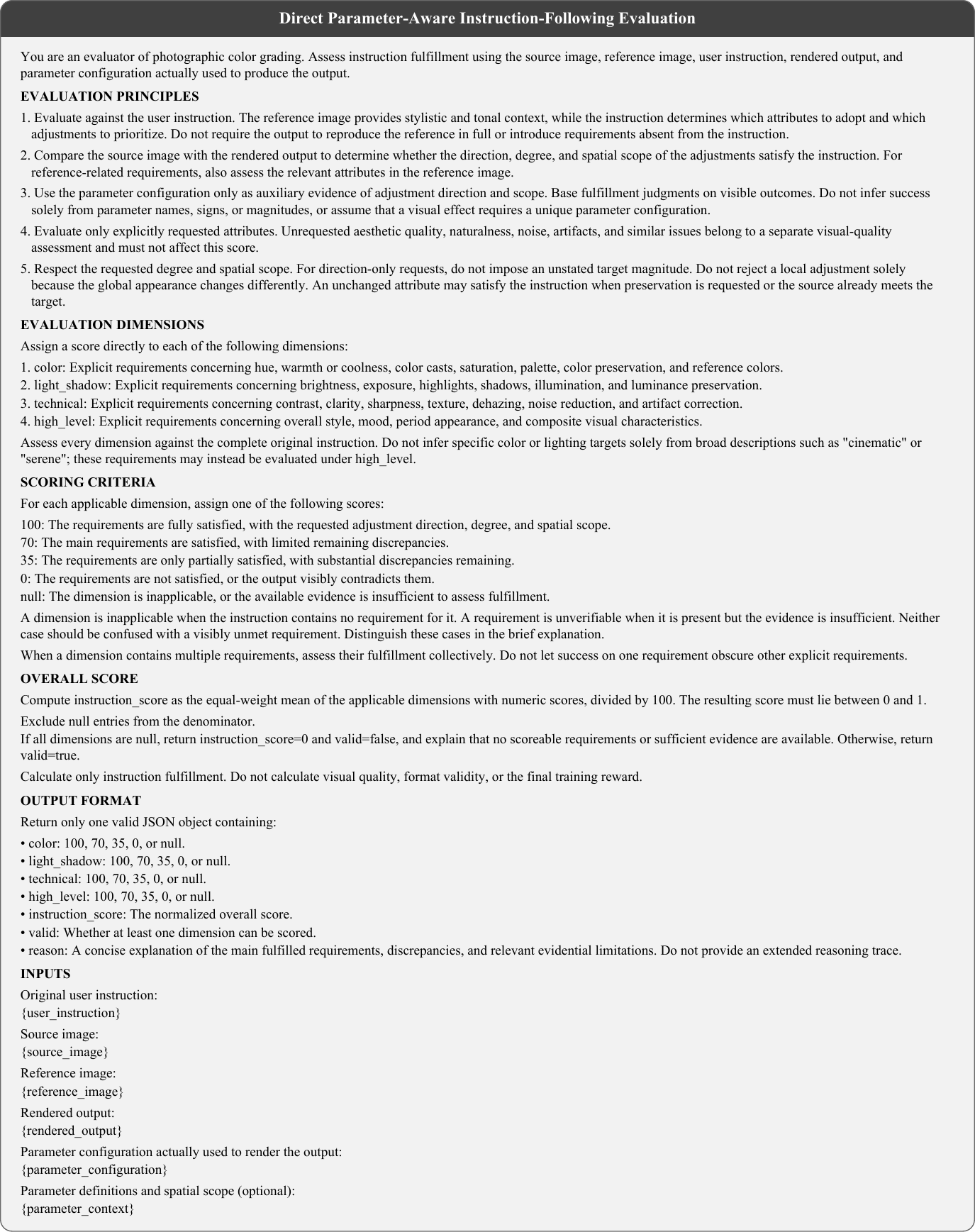}
    \caption{Prompt template for parameter-aware instruction-following reward evaluation.}
    \label{fig:supp-prompt-task-reward}
\end{figure*}

The evaluator receives only the source image and the rendered output, without access to the reference image, user instruction, or executed parameter configuration. Following the prompt in Figure~\ref{fig:supp-prompt-quality-reward}, the evaluator is invoked independently once for each criterion, covering tonal integrity, color fidelity, edge and material fidelity, geometric integrity, and residual artifacts (Table~\ref{tab:supp-quality-criteria}). Each criterion receives a score of $100$, $70$, $35$, or $0$, corresponding to different levels of degradation, or \texttt{null} when no relevant subject is present (Table~\ref{tab:supp-quality-levels}). Applicable scores must be supported by concrete comparisons between the source and rendered images. An inapplicable criterion incurs no penalty and is assigned a score of $100$ during aggregation.

The evaluator does not compute an aggregate score. Instead, the visual-quality reward is calculated externally as
\begin{equation}
  R_i^{\mathrm{qual}}=\frac{1}{100K}\sum_{k=1}^{K}q_{i,k},
  \qquad K=14,
\end{equation}
where $q_{i,k}\in\{0,35,70,100\}$ and \texttt{null} outputs are mapped to $100$, yielding $R_i^{\mathrm{qual}}\in[0,1]$.

\begin{table*}[t]
  \caption{Four-level scoring scheme for visual quality.}
  \label{tab:supp-quality-levels}
  \centering
  \begin{tabularx}{\linewidth}{@{}lcX@{}}
    \toprule
    Level & Numerical score & Definition \\
    \midrule
    Level 1 & 100 & The property is visibly preserved from the source, with a concrete supporting observation. \\
    Level 2 & 70 & A trace or mild defect is visible only on close inspection, or the property cannot be conclusively judged from the supplied images. \\
    Level 3 & 35 & A defect is clearly visible under normal viewing and materially affects the property. \\
    Level 4 & 0 & A severe or critical defect makes the property unacceptable. \\
    \bottomrule
  \end{tabularx}
\end{table*}

\subsection{Format-Validity Reward}
\label{sec:supp-format-reward}

The format-validity reward has two independently evaluated parts. The response is written as \texttt{<think>...</think>} followed by \texttt{<tool\_call>}$P$\texttt{</tool\_call>}.

\begin{enumerate}
\item \textbf{Response and XML parsing.}
Let $F$ denote the parsing score.  It is the sum of four independent binary checks:
\begin{equation}
  F = 0.2 I_{\mathrm{think}}
    + 0.2 I_{\mathrm{tool\_call}}
    + 0.4 I_{\mathrm{xml}}
    + 0.2 I_{\mathrm{root}}.
\end{equation}
The reasoning and tool-call tag pairs must each occur exactly once, with the opening tag
before the closing tag.  The XML declaration is removed before parsing.  The content inside
\texttt{tool\_call} is parsed with the project's sign-aware XML converter, and its root
element must be literally \texttt{root}.  A failed sub-check lowers $F$ without necessarily
making the entire parsing score zero.

\item \textbf{Parameter validation.}
Let $P$ denote the parameter-validation score.  Missing or empty parameters, a fatal
structural or type error, or a failed parameter-structure check yields $P=0$ and
\texttt{valid}=0.  Otherwise, the validator checks registered enumerations, the nested
structures handled explicitly by the implementation, and registered top-level numeric
fields.  For a registered numeric parameter with legal interval $[l,u]$ and value $v$,
define the normalized excess
\begin{equation}
  e(v;l,u)=\frac{\max\{0,\,l-v,\,v-u\}}{u-l}.
\end{equation}
The corresponding numeric score is $1.0$ when $e=0$, $0.9$ when $0<e\leq 0.1$,
$0.7$ when $0.1<e\leq 0.5$, and $0.2$ when $e>0.5$.  Numeric scores are averaged
within each populated category and combined with the implementation's category weights
\{basic:0.40, color:0.20, detail:0.10, advanced:0.05, mask:0.25\}; an empty category
defaults to $1.0$.

Let $\mathcal{S}$ be the supported top-level parameter schema and let $\mathcal{K}$ be
the set of top-level keys in the output.  Define
$\mathcal{U}=\{k\mid k\in\mathcal{K},\,k\notin\mathcal{S}\}$.
Unknown names are treated as anomalous output: define the name gate
$I_{\mathrm{name}}=\mathbf{1}[\mathcal{U}=\varnothing]$.  Likewise, let
$I_{\mathrm{struct}}$ indicate that the mapping is non-empty and passes all registered
enum, nested-structure, and type checks.  The parameter-validation score is then defined as
\begin{equation}
  P=I_{\mathrm{name}}\,I_{\mathrm{struct}}\,R_{\mathrm{range}},
\end{equation}
where $R_{\mathrm{range}}$ is the category-weighted score from the registered numeric
range checks above.  Thus, any unknown parameter name or fatal structural error gives
$P=0$, while a registered numeric value outside its range can retain graded partial credit.
\end{enumerate}

Table~\ref{tab:supp-format-checks} summarizes both parts of the format-validity reward. 
For candidate $i$, the format-validity reward is
\begin{equation}
  R_i^{\mathrm{fmt}} = P\,F,
\end{equation}
where $P$ and $F$ are the parameter-validation and parsing scores for that candidate, respectively.

\section{Additional Experimental Details}
\label{sec:supp-exp-details}

\subsection{Detailed Experimental Setup}
\label{sec:supp-training}

JarvisAura is developed from Qwen3-VL-8B-Instruct through PG-CPT, RG-SFT, and RA-GRPO.
Table~\ref{tab:supp-hyperparams} summarizes the three-stage training configuration.

\subsection{Construction of Ref-Bench}
\label{sec:supp-benchmark}
Ref-Bench comprises 300 test instances and follows the same data collection, parameterized expert-editing, and consensus-validation protocol as the approximately 2K-sample Ref-RL dataset. The two datasets are independently collected and have no overlap in users, source images, user instructions, expert-edited targets, or reference images. In particular, Ref-Bench uses a separate reference-image library, and its expert editors are distinct from those involved in Ref-RL construction. Its samples further cover broader ranges of user ages, imaging expertise, and photographic scenes. Thus, Ref-Bench evaluates generalization to unseen cases within the same task domain.

\subsection{Automatic Evaluation Protocols}
\label{sec:supp-automatic-evaluation}

\noindent\textbf{Target-based fidelity.}
To measure how closely each edited image agrees with the corresponding expert consensus result, we compare it with the expert-refined target.
The edited image and target are independently center-cropped to a square using the shortest image side and then resized to $512\times512$ pixels with Lanczos interpolation.
RGB values are represented in normalized sRGB, with values in $[0,1]$.
For an edited image $I$ and target $T$, L1 is the mean absolute RGB error over all pixels and channels,
\begin{equation}
  \mathrm{L1}(I,T)
  = \frac{1}{3HW}
    \sum_{x=1}^{W}\sum_{y=1}^{H}\sum_{c\in\{R,G,B\}}
    \left|I_{x,y,c}-T_{x,y,c}\right|,
\end{equation}
where $H=W=512$.
L2 denotes the mean squared error (MSE), rather than the root mean squared error,
\begin{equation}
\begin{aligned}
  \mathrm{L2}(I,T)
  &= \mathrm{MSE}(I,T) \\
  &= \frac{1}{3HW}
     \sum_{x=1}^{W}\sum_{y=1}^{H}\sum_{c\in\{R,G,B\}}
     \left(I_{x,y,c}-T_{x,y,c}\right)^2.
\end{aligned}
\end{equation}
CIEDE2000 measures perceptual color difference.
We convert normalized sRGB values to CIELAB using the D65 reference white and compute the standard per-pixel $\Delta E_{00}$ with $k_L=k_C=k_H=1$:

\begin{equation}
\begin{aligned}
  &\mathrm{CIEDE2000}(I,T) \\
  &\quad= \frac{1}{HW}\sum_{x=1}^{W}\sum_{y=1}^{H}
    \Delta E_{00}\!\left(
      \mathrm{Lab}(I_{x,y}),\mathrm{Lab}(T_{x,y})
    \right).
\end{aligned}
\end{equation}
For each method, we first compute each metric per valid test case and then report the arithmetic mean over valid cases; missing or invalid result images are excluded from aggregation.
Lower values indicate better target fidelity.
Main-text tables report L1 multiplied by $10^2$ and L2 (MSE) multiplied by $10^3$.
Table~\ref{tab:supp-metric-settings} summarizes preprocessing and aggregation settings for these metrics.

\begin{table*}[t]
  \caption{Preprocessing and aggregation settings for automatic metrics.}
  \label{tab:supp-metric-settings}
  \centering
  \begin{tabularx}{\linewidth}{@{}p{0.40\linewidth}X@{}}
    \toprule
    Setting & Configuration \\
    \midrule
    Image resizing and alignment & Independently center-crop the edited image and target to the shortest-side square; resize both to $512\times512$ with Lanczos interpolation \\
    Pixel range and normalization & RGB values represented as normalized sRGB in $[0,1]$ \\
    Color profile and transfer function & IEC 61966-2-1 sRGB transfer function \\
    Lab conversion and reference white & sRGB-to-CIELAB conversion with the D65 reference white \\
    CIEDE2000 implementation and factors & Standard CIEDE2000 formulation with $k_L=k_C=k_H=1$ \\
    Spatial and channel reduction & L1/MSE averaged over all $H\times W\times3$ RGB samples; CIEDE2000 averaged over all $H\times W$ pixels \\
    Dataset-level aggregation & Arithmetic mean over valid test cases \\
    Invalid or missing outputs & Excluded from aggregation \\
    \bottomrule
  \end{tabularx}
\end{table*}

\noindent\textbf{Reference-guided Semantic consistency and Perceptual Quality.}
To evaluate the perceptual quality of reference-guided editing without relying on a ground-truth image, we use Gemini-3.1-Pro to assess each result using the corresponding user instruction and three images provided in a fixed order: the original source image, the style reference image, and the edited image.
The model assigns two independent scores on a 0--10 scale using half-point increments.
Semantic Consistency (SC) measures how faithfully the edited image follows the user instruction by selecting and transferring the requested color and style characteristics from the reference image to the source image, while avoiding unrequested reference attributes. 
While Perceptual Quality (PQ) measures preservation of the source content together with the technical quality of the edited image.
The overall score is computed per sample as the geometric mean $O=\sqrt{\mathrm{SC}\times\mathrm{PQ}}$, after which we aggregate the per-sample scores across the evaluation set.
The evaluation prompts are shown in Figure~\ref{fig:supp-eval}.
Training and evaluation use separate prompts: the former is task-specific and requires configuration parameters, whereas the latter is fixed across methods to ensure fair comparison.

\subsection{Human Evaluation Protocols}
\label{sec:supp-human-evaluation}

Human evaluation comprises task-success assessment and pairwise preference assessment.
Both evaluations are blinded, and expert target images are withheld.
All evaluations are conducted on the same latest-generation MacBook Air under a consistent display setup.

\noindent\textbf{Human-rated task success rate.}
HTSR is measured in a blinded user study with expert target images withheld.
The study involves 30 participants spanning different age groups and levels of imaging expertise, including both imaging professionals and non-professional users.
None of the participants were involved in training-data screening or consensus validation.
Using the same latest-generation MacBook Air, participants independently assign one of three labels: successful, uncertain, or unsuccessful.
No fine-grained scoring rubric is imposed; participants make an overall judgment under a realistic user setting while considering fine-grained instruction fulfillment, alignment with requested reference attributes, visual quality, and content preservation.
The final label is determined by majority vote, and a case is counted as successful only when successful receives the majority.

\begin{table*}[tb]
  \caption{Training time for each stage of JarvisAura, measured in GPU hours on a single A100.}
  \label{tab:supp-training-time}
  \centering
  \setlength{\tabcolsep}{10pt}
  \begin{tabular}{@{}llcc@{}}
    \toprule
    \textbf{Stage} & \textbf{Configuration} & \textbf{Samples} & \textbf{GPU Hours (A100)} \\
    \midrule
    PG-CPT   & 1 epoch  & 200K & $\sim$500 \\
    RG-SFT   & 2 epochs & 50K  & $\sim$160 \\
    RA-GRPO & 1 epoch & 2K & $\sim$ 280 \\
    \midrule
    Total & & & $\sim$940 \\
    \bottomrule
  \end{tabular}
\end{table*}

\noindent\textbf{Pairwise A/B preference.}
Using the same participants and device, JarvisAura is compared separately with GPT-Image-2.5, Seedream-5-pro, Nano-Banana-2, and FireRed-Image-Edit.
Each comparison uses the intersection of cases judged successful for JarvisAura and that baseline, not a common intersection across all methods.
For each case, the outcome is determined by majority vote: JarvisAura is counted as winning only when it receives more votes than the compared baseline, while an equal vote count is recorded as a tie.
The reported preference ratio is
\begin{equation}
  \mathrm{PR}
  =
  \frac{N_{\mathrm{win}}+N_{\mathrm{tie}}}
       {N_{\mathrm{loss}}+N_{\mathrm{tie}}},
  \label{eq:supp-preference-ratio}
\end{equation}
where wins and losses are defined from JarvisAura's perspective.
A value above one favors JarvisAura.
This quantity is a ratio, not a preference percentage or percentile.

\subsection{Training and Inference Time}
\label{sec:supp-computational-cost}

Tables~\ref{tab:supp-training-time} and~\ref{tab:supp-inference-time} summarize training and inference costs, respectively.

\begin{table*}[t]
  \caption{Inference configurations and computational costs.}
  \label{tab:supp-inference-time}
  \centering
  \begin{tabularx}{\linewidth}{@{}Xcc@{}}
    \toprule
    Inference framework & vLLM & Transformers \\
    \midrule
    End-to-end time (s/sample) & 38.18 & 47.75 \\
    Model generation time (s/sample) & 29.36 & 36.24 \\
    Backend rendering time (s/sample) & 8.75 & 11.22 \\
    Time to first token (s) & 4.19 & 1.38 \\
    Throughput (tokens/s) & 71.32 & 29.73 \\
    Generated tokens per sample & 2,094 & 1078 \\
    \bottomrule
  \end{tabularx}
\end{table*}

\section{Additional Qualitative Results}
\label{sec:supp-qualitative}

\subsection{Intermediate Reasoning and Parameter Outputs}
\label{sec:supp-intermediate-outputs}

Figure~\ref{fig:supp-infer} presents source and reference images, the user instruction, generated reasoning, executable parameter configurations, and rendered outcomes.
These examples expose the agent's intermediate outputs and their relation to the resulting edits.

\subsection{More Visual Comparisons with Generative Methods}
\label{sec:supp-visual-comparisons}
Figure~\ref{fig:supp-morevis} presents additional visual comparisons with input-compatible generative methods.
These methods receive the same source image, reference image, and user instruction as JarvisAura, enabling a controlled and fair comparison.
The examples focus on instruction-specified reference attributes, grading accuracy, visual quality, and source-content preservation.

\begin{table*}[t]
  \caption{Reference-inspection checklist for Ref-SFT quality control.}
  \label{tab:supp-inspection-guide}
  \centering
  \renewcommand{\arraystretch}{1.1}
  \begin{tabularx}{\linewidth}{@{}p{0.27\linewidth}X@{}}
    \toprule
    Dimension & Inspection focus \\
    \midrule
    Semantic correspondence & Correspondence of relevant content such as sky, skin, vegetation, water, and architectural regions. \\
    Overall color tone & Dominant color tendencies, white balance, and overall color coherence. \\
    Warm--cool balance & Relationships among highlights, midtones, and shadows. \\
    Lighting atmosphere & Compatibility of the reference's lighting appearance with the intended grading changes. \\
    Tonal structure and contrast & Highlight transitions, shadow detail, midtone gradation, and contrast. \\
    Local texture and clarity & Relevant textural characteristics without conflating grading with content replacement. \\
    Subject--background separation & Tonal and color relationships contributing to foreground and background separation. \\
    Style coherence & Consistency of the reference attributes with the intended grading direction. \\
    \bottomrule
  \end{tabularx}
\end{table*}

\begin{table*}[t]
  \caption{Format-validity reward checks.}
  \label{tab:supp-format-checks}
  \centering
  \begin{tabularx}{\linewidth}{@{}p{0.22\linewidth}p{0.48\linewidth}X@{}}
    \toprule
    Check & Implemented property & Numerical rule \\
    \midrule
    Reasoning block & Exactly one ordered \texttt{<think>} pair & $0.20$ in $F$ \\
    Tool-call block & Exactly one ordered \texttt{<tool\_call>} pair & $0.20$ in $F$ \\
    XML parsing & Inner XML parses with sign-aware \texttt{xml\_to\_dict} & $0.40$ in $F$ \\
    Root tag & Parsed XML root is exactly \texttt{root} & $0.20$ in $F$ \\
    Parameter structure & Non-empty mapping; registered enums and explicitly handled nested structures/types are valid & $P=0$ on fatal failure \\
    Parameter names & Every top-level name must belong to the supported parameter schema & Any unknown name gives $P=0$ \\
    Parameter values & Registered top-level numeric fields in the range schema are compared with their declared ranges & $1.0/0.9/0.7/0.2$ by excess \\
    \bottomrule
  \end{tabularx}
\end{table*}

\begin{table*}[tb]
  \caption{Hyperparameters for the three training stages of JarvisAura.}
  \label{tab:supp-hyperparams}
  \centering
  \begin{tabular}{@{}lccc@{}}
    \toprule
    \textbf{Hyperparameter} & \textbf{PG-CPT} & \textbf{RG-SFT} & \textbf{RA-GRPO} \\
    \midrule
    \multicolumn{4}{@{}l}{\textit{Model \& Data Configuration}} \\
    Max Sequence Length        & 24576 & 24576 & 24576 \\
    Train Batch Size (per GPU) & 1     & 1     & 1     \\
    Training Samples           & 200K  & 50K   & 2K + derived samples \\
    Original samples yielding a derived sample & -- & -- & 28\% \\
    \midrule
    \multicolumn{4}{@{}l}{\textit{Optimizer \& Scheduler}} \\
    Optimizer                  & AdamW & AdamW & AdamW \\
    Learning Rate              & $5\times10^{-5}$ & $1\times10^{-5}$ & $5\times10^{-6}$ \\
    LR Scheduler               & Cosine & Cosine & Constant \\
    Warmup Ratio               & 0.1   & 0.1   & 0     \\
    \midrule
    \multicolumn{4}{@{}l}{\textit{Training Details}} \\
    Training Epochs            & 1     & 2     & 1     \\
    Precision                  & bf16  & bf16  & bf16  \\
    \midrule
    \multicolumn{4}{@{}l}{\textit{RA-GRPO Configuration}} \\
    Max Prompt Length          & --    & --    & 17900 \\
    Max Response Length        & --    & --    & 6384  \\
    Advantage Estimator         & --    & --    & GRPO  \\
    Rollout Backend             & --    & --    & vLLM (async) \\
    Rollouts per Prompt        & --    & --    & 8     \\
    Sampling Temperature       & --    & --    & 0.75  \\
    Sampling Top-$p$           & --    & --    & 0.92  \\
    GPU Count                  & 8$\times$A100 & 8$\times$A100 & 8$\times$A100 \\
    \bottomrule
  \end{tabular}
\end{table*}

\begin{table*}[t]
  \centering
  \caption{Criteria for the visual-quality reward.}
  \label{tab:supp-quality-criteria}
  \setlength{\tabcolsep}{5pt}
  \renewcommand{\arraystretch}{1.0}

  \begin{tabularx}{\textwidth}{
    @{}c
    >{\raggedright\arraybackslash}p{0.25\textwidth}
    >{\raggedright\arraybackslash}X@{}
  }
    \toprule
    Criterion & Name & Definition and defect condition \\
    \midrule
  1 & Highlight and overexposure integrity & Highlights retain roll-off, headroom, and texture relative to the source; broad near-white crowding or a blown key subject is a defect. \\
  2 & Shadow and dark-detail integrity & Shadows retain boundaries, local contrast, and material separation; newly flattened, featureless dark regions are a defect. \\
  3 & Contrast, dynamic-range, and veil integrity & Global and local contrast and tonal separation remain credible; a broad white or colored veil or harsh tonal compression is a defect. \\
  4 & Subject-environment tonal integration & A key subject's brightness and local contrast remain consistent with its surrounding scene; an implausible spotlighted or cutout appearance is a defect. \\
  5 & Gradient, banding, and posterization integrity & Smooth or gently graded regions (sky, water, walls) remain continuous; visible banding, stepped contours, or posterization is a defect. \\
  6 & Local chroma continuity and blocking & Color within a single material or semantic region stays spatially continuous; abrupt within-region color steps or blocky chroma cells are a defect. \\
  7 & Edge, mask, and transition integrity & Boundaries between subjects and regions remain natural; visible halos, color fringing, mask seams, or a cutout appearance are a defect. \\
  8 & Detail and texture integrity & Fine surface detail and micro-contrast (skin, fabric, foliage) are preserved; waxy, over-smoothed, or flattened texture is a defect. \\
  9 & Material rendering fidelity & Expected material cues (specularity, transparency, grain, moisture) remain credible; loss of a material's characteristic surface response is a defect. \\
  10 & Human face and skin chroma naturalness & Visible face, neck, or hand skin tones remain plausible and internally consistent; patchy, contaminated, or implausible skin chroma is a defect. \\
  11 & Neutral and monochrome chromatic integrity & Neutral, gray, or black-and-white regions keep a plausible, spatially consistent color identity; an unexplained localized color cast is a defect. \\
  12 & Non-human subject color plausibility & Colors of food, vegetation, sky, water, buildings, or objects remain semantically and physically plausible; fluorescent or identity-changing color shifts are a defect. \\
  13 & Geometry and semantic structure integrity & Shapes, boundaries, and the recognizability of faces, bodies, and objects are preserved; new warping, merging, or structural corruption is a defect. \\
  14 & Residual processing cleanliness & No newly introduced noise, over-sharpening, aliasing, or compression artifact appears beyond what the other thirteen criteria already describe. \\
    \bottomrule
  \end{tabularx}
\end{table*}

\clearpage
\clearpage

\clearpage
\clearpage

\clearpage
\clearpage

\clearpage
\clearpage

\clearpage

\clearpage

\clearpage

\clearpage

\clearpage
\clearpage

\clearpage
\clearpage

\clearpage

\clearpage

\clearpage

\clearpage

\clearpage

\clearpage
\begin{figure*}[p]
    \centering
    \includegraphics[
        width=\textwidth,
        height=0.85\textheight,
        keepaspectratio
    ]{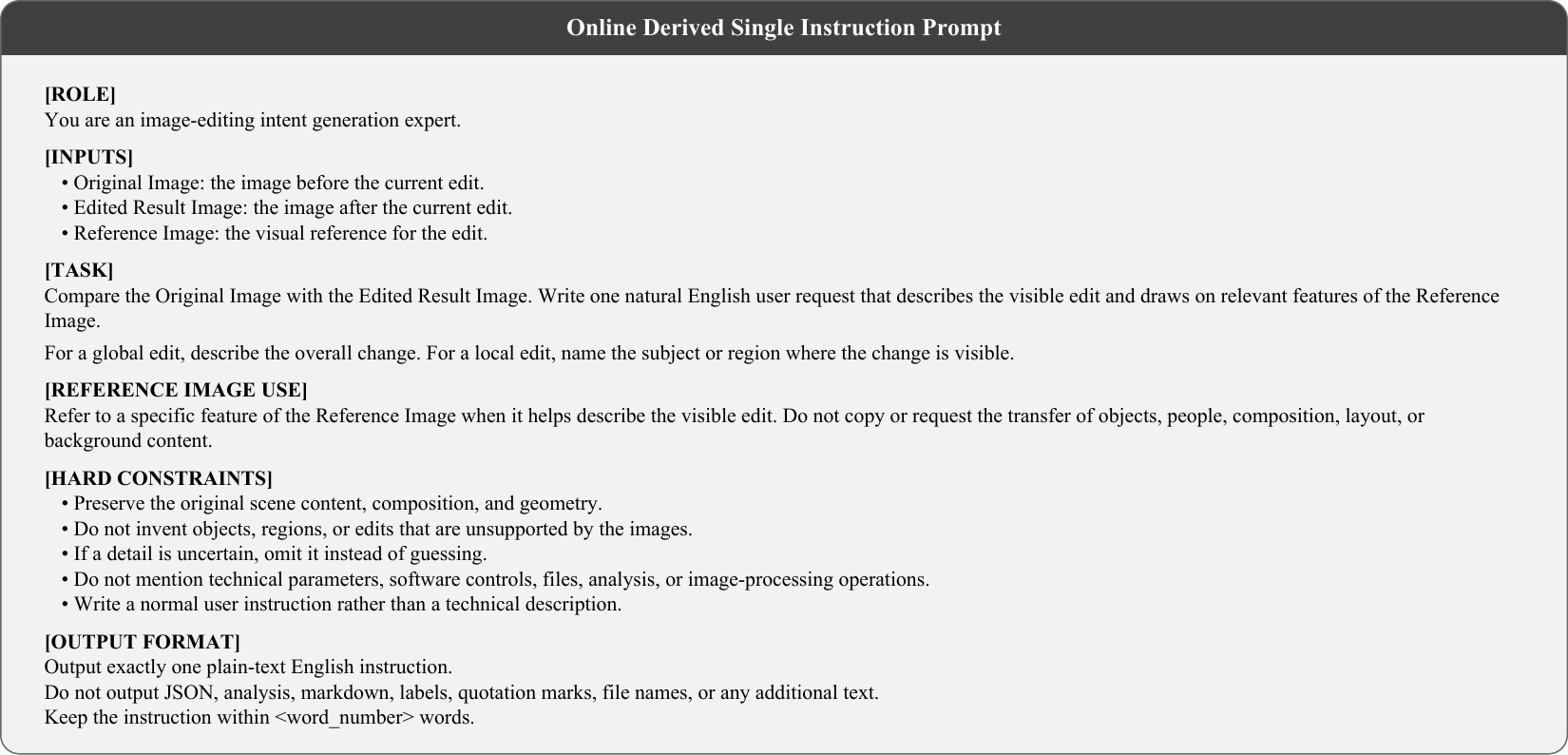}
    \caption{Prompt template for image-conditioned instruction generation during dynamic data derivation.}
    \label{fig:supp-prompt-derivation}
\end{figure*}

\clearpage
\begin{figure*}[p]
    \centering
    \includegraphics[
        width=\textwidth,
        height=0.85\textheight,
        keepaspectratio
    ]{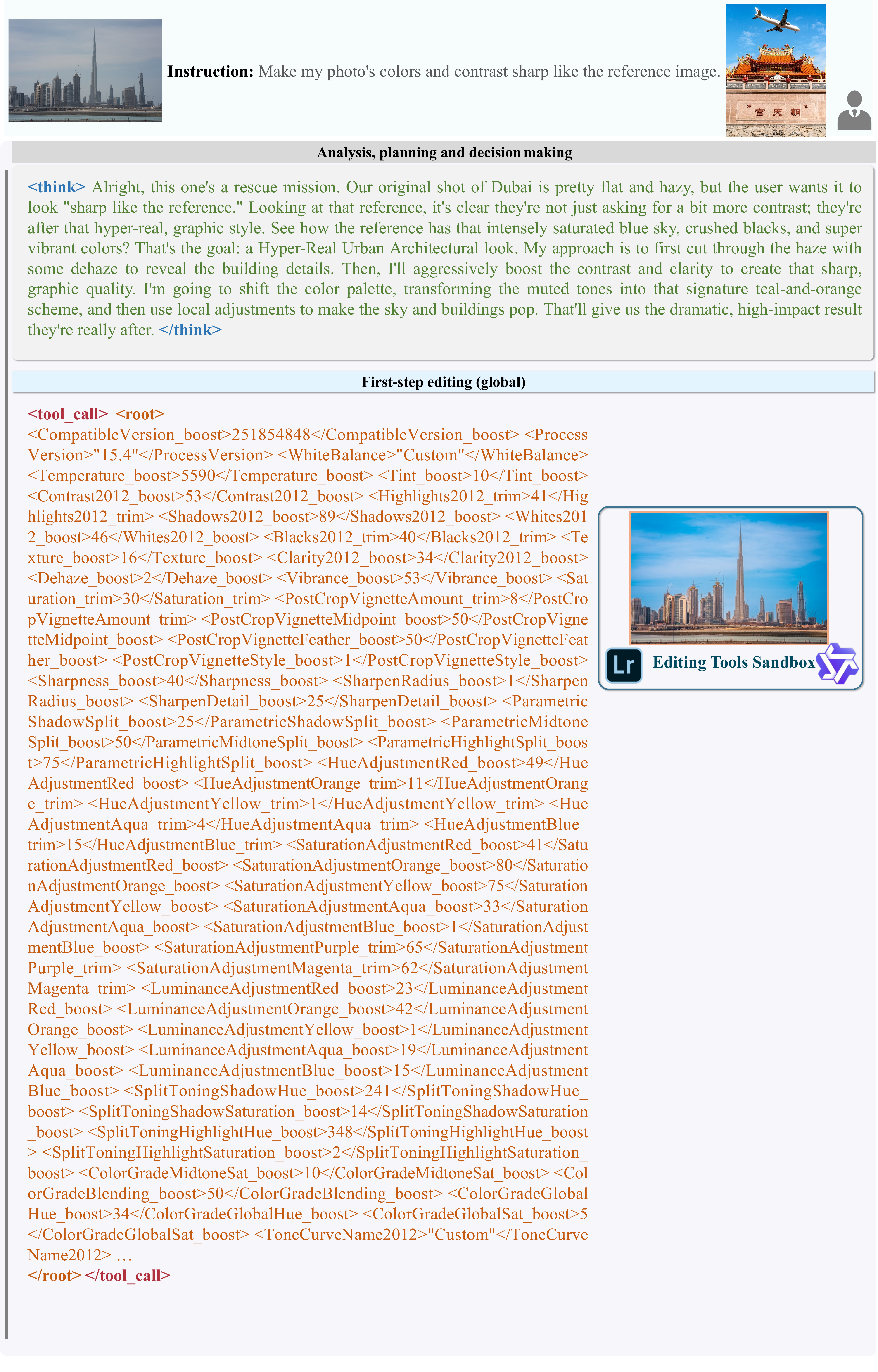}
    \caption{Intermediate reasoning, predicted parameter configurations, and rendered outputs of JarvisAura.}
    \label{fig:supp-infer}
\end{figure*}

\clearpage
\begin{figure*}[p]
    \centering
    \includegraphics[
        width=\textwidth,
        height=0.85\textheight,
        keepaspectratio
    ]{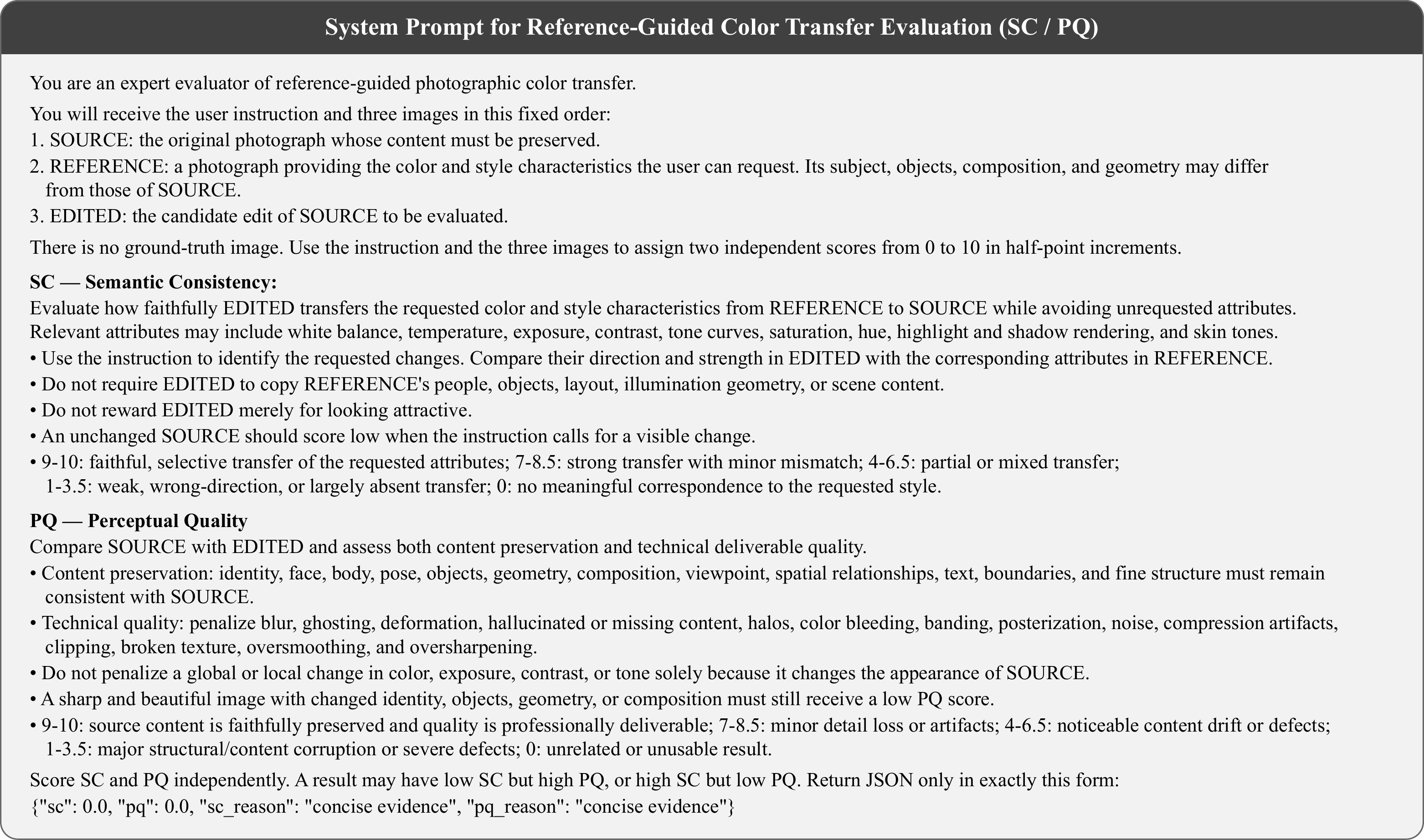}
    \caption{Prompt templates for SC and PQ evaluation on Ref-Bench.}
    \label{fig:supp-eval}
\end{figure*}

\clearpage
\begin{figure*}[p]
    \centering
    \includegraphics[
        width=\textwidth,
        height=0.85\textheight,
        keepaspectratio
    ]{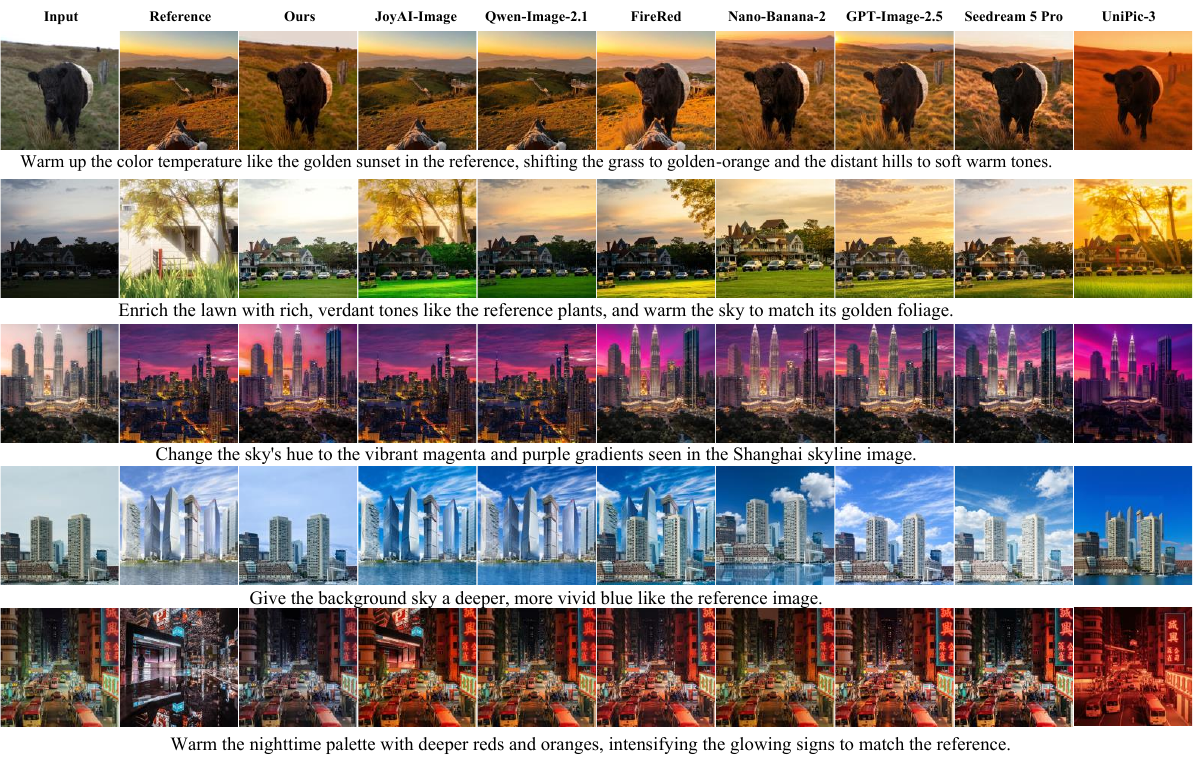}
    \caption{Additional qualitative comparisons on Ref-Bench.}
    \label{fig:supp-morevis}
\end{figure*}
\clearpage

\end{document}